\documentclass[11pt]{article}

\usepackage[preprint]{acl}
\usepackage{times}
\usepackage{latexsym}

\usepackage[T1]{fontenc}

\usepackage[utf8]{inputenc}

\usepackage{microtype}

\usepackage{inconsolata}

\usepackage{graphicx}

\usepackage{amsmath}
\usepackage{amsfonts}   %
\usepackage{multirow}
\usepackage{booktabs}
\usepackage[table]{xcolor}

\usepackage[most]{tcolorbox} %
\newtcolorbox[list inside=CASE, auto counter, number within=section]{CASE}[1][]{
    breakable,
    colbacktitle=black!60,
    coltitle=white,
    fontupper=\footnotesize,
    boxsep=5pt,
    left=0pt,
    right=0pt,
    top=0pt,
    bottom=0pt,
    boxrule=1pt,
    title={#1},
    #1, %
}
\newtcolorbox[list inside=prompt,auto counter,number within=section]{prompt}[1][]{
    colbacktitle=black!60,
    coltitle=white,
    fontupper=\footnotesize,
    boxsep=5pt,
    left=0pt,
    right=0pt,
    top=0pt,
    bottom=0pt,
    boxrule=1pt,
    title={#1},
    #1, %
}

\title{ENTRA: Entropy-Based Redundancy Avoidance in Large Language Model Reasoning}

\author{
  Ruichu Cai$^{1,2}$ \quad
  Haopeng Du$^1$ \quad
  Qingwen Lin$^{1}$ \quad
  Yutong Chen$^{1}$ \quad
  Zijian Li$^{3,4}$ \quad 
  Boyan Xu$^{1}$\thanks{\ \ Corresponding author, \url{hpakyim@gmail.com}} \\
  $^1$School of Computer Science, Guangdong University of Technology \\
  $^2$Peng Cheng Laboratory \quad 
  $^3$Carnegie Mellon University \\
  $^4$Mohamed bin Zayed University of Artificial Intelligence\\
  \texttt{cairuichu@gmail.com, dhp975751379@gmail.com, qingwen\_lin@foxmail.com} \\
  \texttt{\{babbowabby, leizigin, hpakyim\}@gmail.com}
}

\begin{document}
\maketitle
\begin{abstract}
Large Reasoning Models (LRMs) often suffer from \textit{overthinking}, generating unnecessarily long reasoning chains even for simple tasks. This leads to substantial computational overhead with limited performance gain, primarily due to redundant verification and repetitive generation. While prior work typically constrains output length or optimizes correctness, such coarse supervision fails to guide models toward concise yet accurate inference.
In this paper, we propose \textbf{ENTRA}, an entropy-based training framework that suppresses redundant reasoning while preserving performance. ENTRA first estimates the token-level importance using a lightweight \textbf{Bidirectional Importance Estimation (BIE)} method, which accounts for both prediction confidence and forward influence. It then computes a redundancy reward based on the entropy of low-importance tokens, normalized by its theoretical upper bound, and optimizes this reward via reinforcement learning.
Experiments on mathematical reasoning benchmarks demonstrate that ENTRA reduces output length by 37\% to 53\% with no loss—and in some cases, gains—in accuracy. Our approach offers a principled and efficient solution to reduce overthinking in LRMs, and provides a generalizable path toward redundancy-aware reasoning optimization.
\end{abstract}

\section{Introduction}
\begin{figure}[t]
\centering
\includegraphics[width=1\columnwidth]{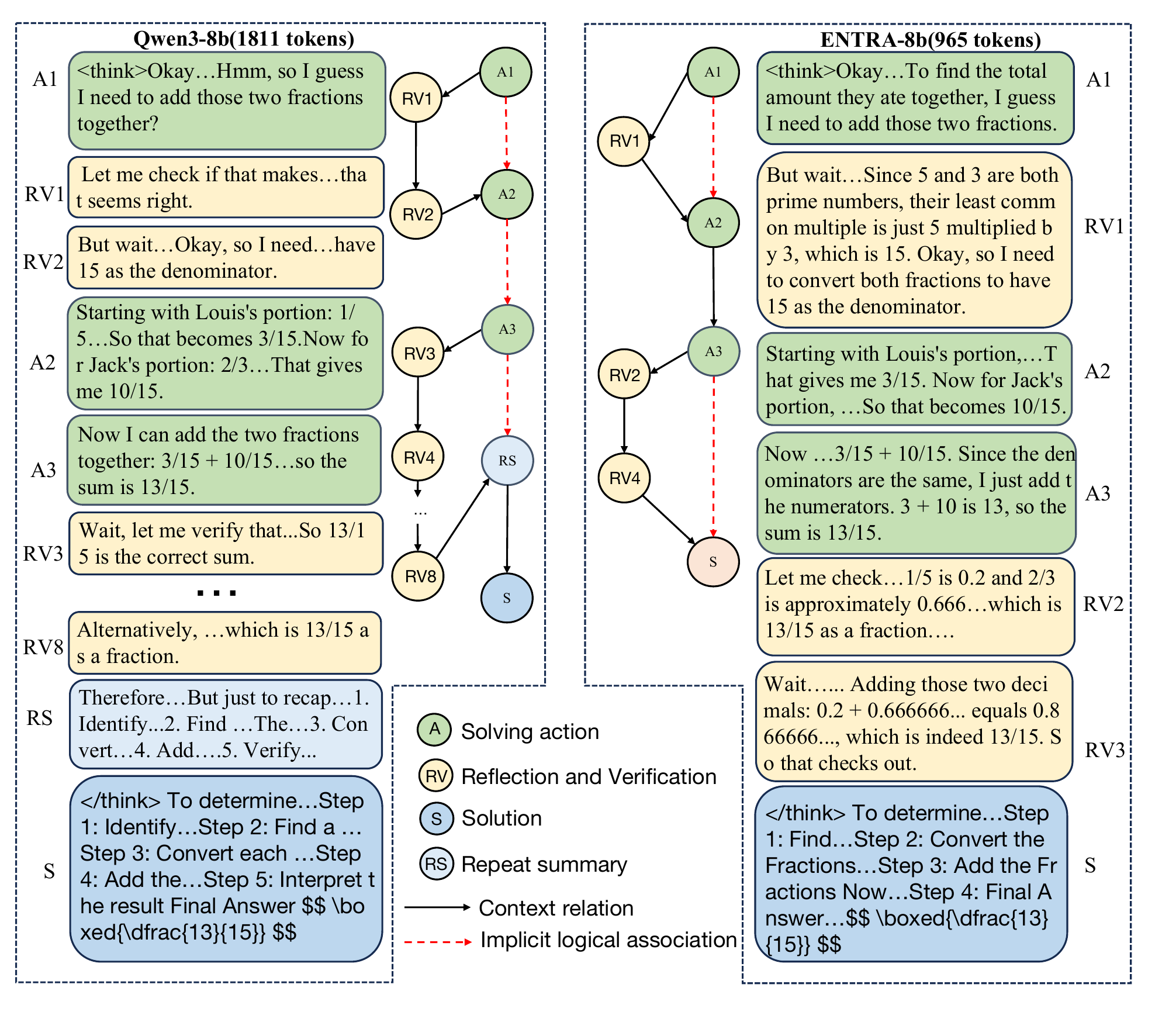} 
\caption{After training on the basis of Qwen3-8b, our method can effectively reduce the steps of excessive inference and guarantee the correct answer.}
\label{fig1}
\end{figure}

Recent advances in Chain-of-Thought (CoT) prompting~\cite{wei2023chainofthoughtpromptingelicitsreasoning} have significantly enhanced the reasoning capabilities of Large Language Models (LLMs). Compared to traditional single-step or shallow reasoning approaches, CoT encourages models to generate intermediate steps that reflect structured thinking. Building upon this idea, models such as OpenAI's o1, DeepSeek-R1~\cite{deepseekai2025deepseekr1incentivizingreasoningcapability}, and Alibaba's QwQ~\cite{qwq32b} introduce mechanisms for self-reflection, multi-path exploration, and answer verification. These techniques extend short CoT into long-form reasoning processes that improve accuracy and robustness. We refer to such models with extended reasoning behaviors as \textit{Large Reasoning Models (LRMs)}~\cite{xu2025largereasoningmodelssurvey}.

However, as LLMs grow in scale and are applied to increasingly diverse tasks, LRMs have begun to exhibit a notable drawback: \textit{overthinking}~\cite{qu2025surveyefficientreasoninglarge,cuadron2025dangeroverthinkingexaminingreasoningaction}. Even when facing simple queries, these models tend to produce excessively long and redundant reasoning chains. For example, when answering “What is 2 + 3?”, QwQ-32B generates over 900 tokens and explores more than ten different solution paths before arriving at the correct answer~\cite{chen2025think23overthinkingo1like}. While the final answer may still be correct, the computation involved is disproportionately large relative to the task's simplicity.
Such redundant reasoning manifests as repeated intermediate steps, unnecessary verification, or over-decomposition of trivial subproblems. This not only affects readability and efficiency but also significantly increases inference cost—especially given the quadratic attention complexity of Transformer-based architectures~\cite{vaswani2017attention} and their reliance on KV cache. Existing efforts in efficient reasoning mostly focus on using output length or correctness as supervision signals to shorten responses. However, these coarse metrics fail to distinguish between informative reasoning and superficial verbosity, and often lead to unstable or suboptimal outcomes. As illustrated in Figure~\ref{fig1}, a typical LRM (Qwen3-8B) generates an extended reasoning trace involving numerous reflection and verification steps—even for elementary fraction addition.

To address the overthinking phenomenon in LRMs, we propose \textbf{ENTRA}, a unified training framework that reduces repetitive reasoning while preserving task accuracy. ENTRA consists of three key components: (1) an unsupervised token-level importance estimator to identify critical reasoning steps; (2) a redundancy measurement module based on entropy over low-importance tokens; and (3) a reinforcement learning optimization strategy that minimizes redundancy under theoretical entropy bounds.
As illustrated in Figure~\ref{fig1}, ENTRA significantly suppresses repeated verification and reflection steps in LRM outputs, generating shorter and more interpretable reasoning chains without compromising correctness. Redundancy is formalized as \textit{repetitive generation}, often characterized by excessive reuse of non-essential tokens. To avoid penalizing informative content, we first estimate each token’s bidirectional importance using our \textbf{BIE} method, which combines left-to-right confidence with attention-based forward contribution.
Based on these scores, we compute the entropy over low-importance tokens and normalize it using a theoretical upper bound to form a stable reward function. This reward guides fine-tuning via reinforcement learning, enabling the model to dynamically prune overthinking behaviors. 

\noindent\textbf{Our contributions are summarized as follows:}
\begin{itemize}
    \item We propose \textbf{ENTRA}, a unified training framework for reducing repetitive reasoning in Large Reasoning Models. ENTRA integrates token-level importance estimation, redundancy-aware reward construction, and reinforcement learning into a coherent optimization pipeline.
    
    \item We introduce \textbf{BIE} (Bidirectional Importance Estimation), a lightweight, training-free method to evaluate token significance based on both prediction confidence and forward influence, enabling selective exclusion of critical tokens from redundancy evaluation.
    
    \item We design a reward function based on entropy over low-importance tokens, normalized by its theoretical upper bound. When optimized via GRPO, our method reduces response length by 37–53\% on benchmark datasets, while maintaining or even improving answer accuracy.
\end{itemize}

\section{Related Work}
In this section, we begin our discussion with the current state of reasoning compression in LLMs and the issue of redundancy. First, we review the phenomenon of “overthinking” in LRMs and typical CoT compression methods in the field of reinforcement learning. Finally, we summarize representative existing works on token importance-based compression.

\subsection{Overthinking in Long CoT}
Chain-of-Thought (CoT) prompting~\cite{wei2023chainofthoughtpromptingelicitsreasoning} has significantly improved large language models' (LLMs) ability to solve complex reasoning tasks by encouraging step-by-step inference. Building on this idea, a range of advanced reasoning paradigms—such as Self-Consistency~\cite{wang2023selfconsistencyimproveschainthought}, Tree-of-Thought~\cite{yao2023treethoughtsdeliberateproblem}, and Monte Carlo Tree Search (MCTS)~\cite{zhang2024restmctsllmselftrainingprocess}—have been proposed to expand the model's exploration space, enhance reflection, and refine intermediate reasoning.

These developments have culminated in the emergence of \textit{Long CoT} strategies, which emphasize multi-step, reflective inference to improve accuracy on complex problems~\cite{chen2025reasoningerasurveylong}. Notably, high-performing models such as OpenAI’s o1 and DeepSeek-R1 demonstrate that longer reasoning chains can yield stronger performance in tasks like math problem solving and code generation.

However, recent studies~\cite{chen2025think23overthinkingo1like, cuadron2025dangeroverthinkingexaminingreasoningaction} have revealed a growing concern: longer reasoning is not always better. Beyond a certain point, LLMs tend to generate repetitive or tangential steps that offer little value—an issue termed \textit{overthinking}. This phenomenon leads to unnecessary token generation, inflated computational cost, and even degraded answer quality, especially when the model loses focus on core problem constraints.

\subsection{Compression Methods for Overthinking}

To alleviate overthinking, recent work has focused on compressing CoT reasoning to improve efficiency without sacrificing accuracy. One common strategy is to apply reinforcement learning (RL) to guide models toward generating concise yet correct reasoning paths. For example, LCPO~\cite{aggarwal2025l1controllinglongreasoning} introduces a goal-aware length reward based on a reference length target, encouraging models to align with desired output lengths. DAST~\cite{shen2025dastdifficultyadaptiveslowthinkinglarge} proposes a difficulty-aware reward that promotes simpler reasoning for easy questions. HAPO~\cite{huang2025hapotraininglanguagemodels} designs a history-based length controller to jointly optimize for brevity and correctness. While effective to some degree, these methods often rely on coarse-grained signals—such as final correctness or output length—which fail to distinguish necessary reasoning steps from redundant ones.

Another line of research leverages token-level importance to guide compression. Selective Context~\cite{li2023compressingcontextenhanceinference} measures self-information to prune uninformative input tokens during prompting. LLMLingua-2~\cite{pan2024llmlingua2datadistillationefficient} formulates prompt compression as a token classification task and distills a lightweight model to preserve semantics while reducing length. Extending this idea to reasoning outputs, TokenSkip~\cite{xia2025tokenskipcontrollablechainofthoughtcompression} assigns token-level importance scores to CoT responses and removes low-contribution tokens under different compression ratios.

While prior methods have made progress in reducing reasoning length, they often struggle to effectively distinguish meaningful reasoning from superficial verbosity. Length-based rewards, though easy to implement, are coarse and may inadvertently suppress necessary intermediate steps. Similarly, correctness-based objectives focus only on the final answer, offering limited guidance on the reasoning process itself.

Token-level compression techniques improve granularity but frequently rely on static, offline scoring mechanisms or external classifiers. These approaches are difficult to generalize across models or tasks and are rarely optimized in an end-to-end training setting. Moreover, few existing methods explicitly quantify the informational redundancy introduced by overthinking, or provide principled constraints to prevent excessive filtering of useful content.

To address these limitations, we advocate for a process-level view of redundancy grounded in information theory. Instead of relying solely on heuristics or black-box scores, our method introduces a lightweight, model-internal token importance estimator, combined with an entropy-based redundancy signal. This allows the model to actively identify and suppress unnecessary computation while preserving key reasoning paths.
\section{Redundancy in LLM Reasoning}
\subsection{Redundancy Analysis}
Large Reasoning Models (LRMs), especially those adopting extended Chain-of-Thought (CoT) reasoning, often exhibit \textit{overthinking} — generating overly long, repetitive inference paths even for simple tasks, which adds computational overhead and impairs reasoning interpretability and efficiency.

As shown in Figure 1, upon examining model outputs across tasks, we observe a dominant forms of redundancy: \textbf{redundant verification steps}, where the model re-checks or reiterates conclusions already logically deduced. Taking Qwen3-8B in Figure 1 as an example, when performing reasoning tasks, the model generates a large number of reflective texts and verification steps with high structural and semantic similarity, exhibiting a distinct characteristic of high repetition rate. These patterns are particularly prevalent in long-form reasoning, where models traverse multiple solution paths or excessively validate intermediate results.

To quantify the redundancy within such outputs, we draw upon concepts from information theory — particularly, \textbf{textual entropy}. Entropy serves as a statistical measure of content diversity. For sequences generated by a sufficiently strong model, those exhibiting higher redundancy typically possess lower entropy, as their token distributions become more skewed and repetitive.

\subsection{Problem Defnition}
To mitigate the inefficiencies caused by redundant reasoning, we aim to formalize the problem of \textit{redundancy suppression} in LRM outputs. Specifically, given a prompt $X$, the model generates a response sequence $Y = [y_1, y_2, ..., y_{|Y|}]$. The combined text $T = [X, Y]$ may contain repeated or low-information tokens, especially in extended reasoning steps. We assume $T$ is a valid natural language sentence — i.e., the sequence of tokens obeys the statistical properties of natural language, including Zipf's Law (heavy-tail frequency), local coherence, and syntactic fluency.

Formally, let $T$ consist of $K$ unique token $\{t_1^k, t_2^k, ..., t_K^k\}$, where $|T| \geq K$, and $p(t_i^k)$ denotes the empirical frequency of token $t_i^k$ in $T$. The entropy of the sequence is defined as:
\begin{equation}
H_{X,Y} = -\sum_{i=1}^{K} p(t_i^k) \log p(t_i^k).
\end{equation}

Our optimization objective, then, is to maximize the entropy of model outputs — thereby encouraging diversity and reducing unnecessary repetition. Given model $M_\theta$, we define the goal as:
\begin{equation}
\max_{\theta}  \quad \mathbb{E}_{X \sim M_\theta} [ H_{X,Y} ].
\end{equation}

This formulation provides a principled foundation for our reinforcement learning objective, which we will describe in Section Method.

\section{Method}
In this section, we introduce our proposed method, \textbf{ENTRA}, which aims to reduce redundant reasoning in long-form outputs. Our approach consists of two main components: (1) an unsupervised token-level importance estimation method that identifies semantically redundant tokens, and (2) an entropy-based reward function that guides the model to minimize unnecessary repetition through reinforcement learning. By integrating token importance with contextual entropy, ENTRA encourages more concise and informative CoT generation.

\subsection{Bidirectional Importance Estimation}
In natural language generation and especially mathematical reasoning, each token serves not only as a continuation of prior context but also as a foundation for future predictions. For instance, intermediate values in a math derivation may appear predictable given the left context, yet they are semantically critical for subsequent steps. Hence, to evaluate the true importance of a token, it is insufficient to consider only its left-to-right predictability. We instead propose to assess token importance from a global perspective, capturing both its current confidence and its influence on future content.

Formally, we define the importance of a token $x_i$ as the log-probability of the token given its full context:
\begin{equation}
    I(x_i) = \log P(x_i \mid x_{<i}, x_{>i}).
\end{equation}
This bidirectional formulation reflects both the forward and backward contributions of the token. Applying Bayes' rule, we decompose this expression as:
\begin{equation}
    P(x_i \mid x_{<i}, x_{>i}) = \frac{P(x_{>i} \mid x_i, x_{<i}) \cdot P(x_i \mid x_{<i})}{P(x_{>i} \mid x_{<i})}.
\end{equation}
Taking logarithms on both sides, we derive:
\begin{align}
    I(x_i) &= [\log P(x_{>i} \mid x_i, x_{<i}) - \log P(x_{>i} \mid x_{<i})] \nonumber \\
    &\quad + \log P(x_i \mid x_{<i}).
\end{align}
The second term corresponds to the standard left-to-right prediction confidence of $x_i$, and can be directly computed from the model's output probabilities. However, the first term—representing the influence of $x_i$ on future predictions—requires expensive marginal computations over the future context, which is impractical during training.

To approximate this future influence efficiently, we turn to the attention mechanism in transformer-based LLMs. Specifically, in an autoregressive setting, future tokens cannot attend to $x_i$, but their attention weights \textit{toward} $x_i$ can serve as a proxy for how much $x_i$ contributes to their generation. For a sequence of length $n$, let $A$ denote the attention matrix (excluding self-attention), and consider the $i$-th column representing token $x_i$. We compute the average attention it receives from future tokens:
\begin{equation}
    \mu_{i} = \frac{1}{n-i} \sum_{j=i+1}^{n} a_{ji}, \quad a_{ji} \in \{A - I\},
\end{equation}
where $I$ is the identity matrix used to exclude self-attention.

\begin{figure}[t]
\centering
\includegraphics[width=1\columnwidth]{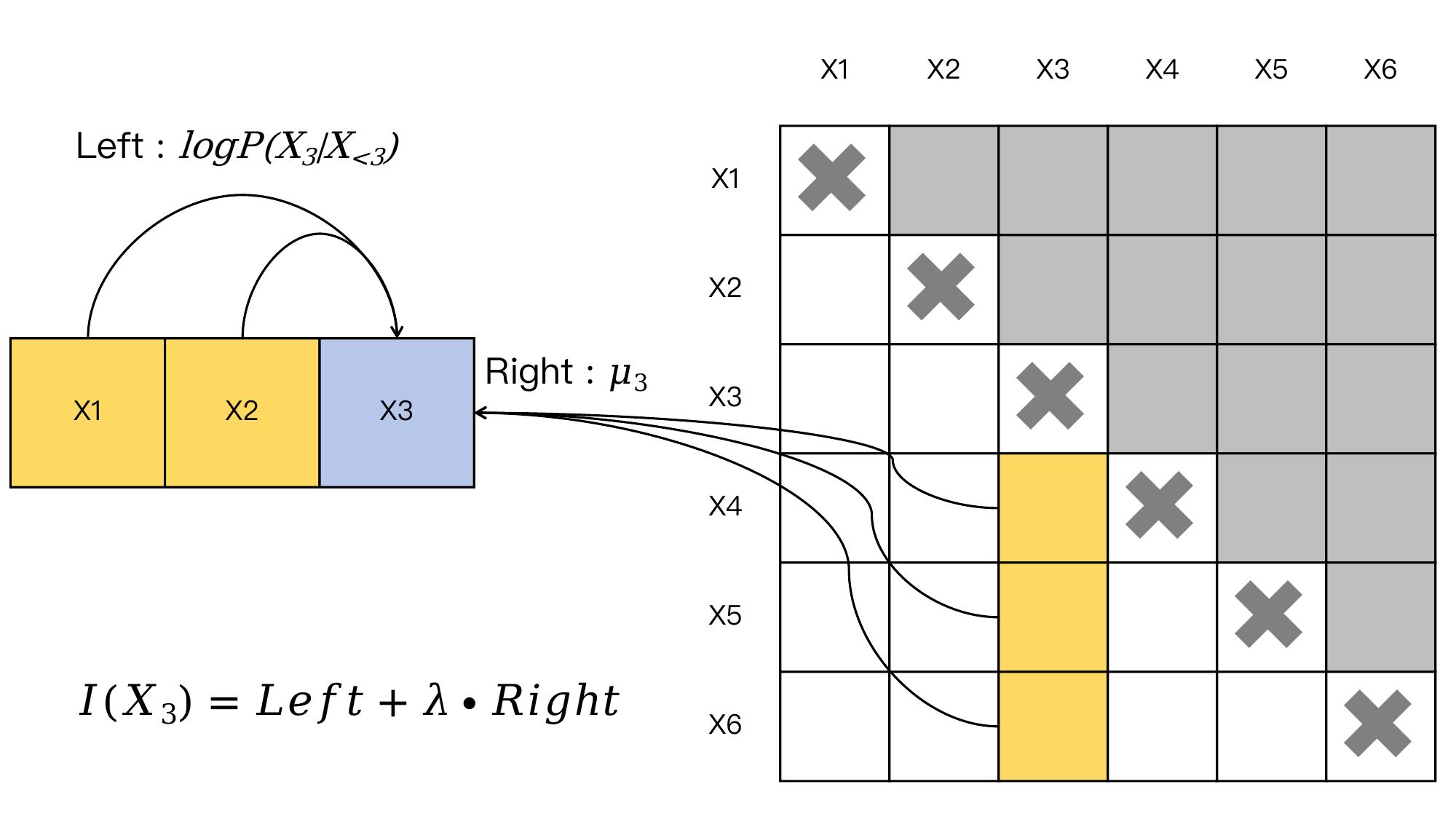} 
\caption{Illustration of our Bidirectional Importance Estimation (BIE). The importance of token $x_3$ is computed by combining its self-information $\log P(x_3 \mid x_{\leq 3})$ (left) with its influence on future tokens, estimated via average attention weights $\mu_3$ (right).}
\label{bie_show}
\end{figure}

We interpret $\mu_i$ as an approximation of $x_i$'s influence on the continuation of the sequence(see Figure~\ref{bie_show} for illustration). Combining this forward influence with the self-information of $x_i$, we define our Bidirectional Importance Estimation (BIE) score as:
\begin{equation}
    I(x_i) \approx \log P(x_i \mid x_{<i}) + \lambda \mu_i,
\end{equation}
where $\lambda$ is a scaling coefficient that balances the two components.

This approximation captures the intuition that important tokens are not only predictable but also influential in shaping subsequent reasoning. In practice, we compute the attention matrix by averaging over all attention heads in the final decoder layer. The hyperparameter $\lambda$ can be tuned per dataset; for instance, we use $\lambda = 2$ on the MATH dataset(See Appendix B for $\lambda$ details).

\subsection{Entropy-Based Redundancy Optimization}
\textbf{From Importance to Redundancy Reward.} 
With the BIE score computed for each token, we identify the subset of tokens with low importance as potential candidates for redundant reasoning. Specifically, we define a threshold $\tau$ and exclude all tokens with $I(x_i) > \tau$ from our redundancy evaluation. The intuition is that highly important tokens—those that are both confidently predicted and highly referenced—should not be penalized during compression.

To quantify the redundancy among the remaining (low-importance) tokens, we turn to a token-level entropy measure, applied only over this filtered subset. This entropy serves as a soft signal of repetition: lower entropy indicates a more skewed, and thus more redundant, token distribution. We use this quantity to define the reward in our reinforcement learning objective.

Let the response text $T$ consist of $K$ unique token types $U = \{t_1, t_2, \dots, t_K\}$, where $|T| \ge K$. Let $u = \{t_i \in U \mid I(t_i) > \tau\}$ be the set of high-importance tokens (above threshold), which are excluded from entropy calculation. The entropy of the remaining tokens is:
\begin{equation}
    H = -\sum_{t_i \in U \setminus u} p(t_i)\log p(t_i).
\end{equation}

Since large language models often generate responses of variable length across different queries—and even across different samples to the same query—the raw entropy values are not directly comparable. This can lead to unstable reward scaling during training. To address this, we normalize the entropy by its theoretical upper bound.

\paragraph{Theoretical Upper Bound of Text Entropy.}
Let a sequence $T$ contain $K$ unique token types, with frequencies $\{p(t_1), p(t_2), ..., p(t_K)\}$ such that $\sum_{i=1}^{K} p(t_i) = 1$. Using the method of Lagrange multipliers to maximize entropy under this constraint, we obtain the maximum entropy when all $p(t_i) = \frac{1}{K}$. That is:
\begin{equation}
    H \le \log K \le \log |T|.
\end{equation}
In our setting, since entropy is computed over the set $U \setminus u$, the final upper bound becomes $\log (|T_i| - k)$, where $k = |u|$ is the number of high-importance tokens.

Thus, for a question $q$ and sampled response set $O = \{o_1, o_2, \dots, o_G\}$, let $T=q+O$, we compute the reward as:
\begin{equation}
    r_i = \frac{-\sum_{t_i \in U \setminus u} p(t_i)\log p(t_i)}{\log (|T_i| - k)}.
\end{equation}

\paragraph{GRPO Optimization.}
We train our model using the GRPO algorithm~\cite{shao2024deepseekmathpushinglimitsmathematical}, a variant of PPO~\cite{2017Proximal} that removes the need for a value network by comparing rewards within sampled groups. This reduces GPU memory usage and improves training stability. For each query $q$, GRPO samples $G$ outputs $O = \{o_1, o_2, \dots, o_G\}$ and optimizes the policy $\pi_\theta$ by maximizing:
\begin{align}
\mathcal{J}_{\text{GRPO}}(\theta) 
&= \mathbb{E}\Big[q \sim P(Q), \{o_i\}_{i=1}^G \sim \pi_{\theta_{\text{old}}}(O|q)\Big] \nonumber\\
&\quad \frac{1}{G} \sum_{i=1}^{G} \Bigg( 
\min\Bigg( 
\frac{\pi_\theta(o_i|q)}{\pi_{\theta_{\text{old}}}(o_i|q)} A_i,\; \nonumber\\
&\quad
\operatorname{clip}\Bigg(
\frac{\pi_\theta(o_i|q)}{\pi_{\theta_{\text{old}}}(o_i|q)}, 1 - \epsilon, 1 + \epsilon
\Bigg) A_i
\Bigg) \nonumber\\
&\quad - \beta \mathbb{D}_{\text{KL}}\Big(\pi_\theta \| \pi_{\text{ref}}\Big)
\Bigg),
\end{align}
where $\epsilon$ and $\beta$ are hyperparameters, and $\mathbb{D}_{\text{KL}}(\cdot\|\cdot)$ denotes the KL divergence between the current model and a reference model. The advantage $A_i$ for each output is computed via intra-group normalization:
\begin{equation}
A_i = \frac{r_i - \operatorname{mean}(\{r_1, r_2, \dots, r_G\})}{\operatorname{std}(\{r_1, r_2, \dots, r_G\})}.
\end{equation}

\begin{table*}[t!]
\centering

\small
\resizebox{\linewidth}{!}{

\begin{tabular}{l|c|cccc|cccc}
\toprule
\multirow{2}{*}{\textbf{Methods}} 
&& \multicolumn{4}{c|}{\textbf{EN}} 
& \multicolumn{4}{c}{\textbf{CN}} \\
 &Result& MATH-500 & GSM8K & MAWPS&AVG 
 & Weekly12K & MATH23K & CMath &AVG\\
\midrule
\midrule
\multicolumn{10}{c}{\textbf{Qwen3-8B}} \\
\midrule
\textbf{Base} &Acc& 0.9460 & 0.9424 & 0.9395&--&0.7990& 0.9630&0.9400&--\\
\midrule
\multirow{2}{*}{\textbf{SFT}}
  &Acc   &0.9480&0.9318&0.9346&$-\phantom{0}0.45\%\,\downarrow$&0.7990&0.9590&0.9383&$-\phantom{0}0.19\%\,\downarrow$\\
  &Compr.&5.41\%&6.94\%&5.50\%&$+\phantom{0}5.95\%\,\uparrow$  &4.84\%&11.94\%&9.24\%&$+\phantom{0}8.67\%\,\uparrow$\\
\midrule
\multirow{2}{*}{\textbf{DPO}}
  &Acc   & 0.9360 & 0.9363 &0.9385 &$-\phantom{0}0.57\%\,\downarrow$&0.7970&0.9600&0.9417&$-\phantom{0}0.11\%\,\downarrow$\\
  &Compr.& 21.96\%& 35.23\%&44.96\%&$+34.05\%\,\uparrow$ &60.37\%&74.05\%&63.57\%&$+\textbf{66.00\%}\,\uparrow$\\
\midrule
\multirow{2}{*}{\textbf{LC-R1}} 
  &Acc   &0.9360 &0.9477&0.9414&$-\phantom{0}0.93\%\,\downarrow$&0.797&0.9620&0.9467&$+\phantom{0}0.12\%\,\uparrow$\\
  &Compr.&27.72\%&53.80\%&56.04\%&$+\textbf{45.85\%}\,\uparrow$&30.20\%&23.72\%&31.32\%&$+28.41\%\,\uparrow$\\
\midrule
\multirow{2}{*}{\textbf{HAPO}} 
  &Acc   &0.9440&0.9462&0.9472&$+\phantom{0}\underline{0.32\%}\,\uparrow$&0.8010&0.9590&0.9483&$+\phantom{0}\underline{0.21\%}\,\uparrow$\\
  &Compr.&11.87\%&3.83\%&7.87\%&$+\phantom{0}7.86\%\,\uparrow$&-12.12\%&-15.85\%&-3.99\%&$-10.65\%\,\downarrow$\\
\midrule
\multirow{2}{*}{\textbf{ENTRA}} 
  &Acc   & 0.9480 & 0.9507 & 0.9429 &$+\phantom{0}\textbf{0.46\%}\,\uparrow$&0.8050&0.9620&0.9483&$+\phantom{0}\textbf{0.44\%}\,\uparrow$\\
  &Compr.& 34.46\%& 43.17\%& 36.01\%&$+\underline{37.88\%}\,\uparrow$&54.05\%&50.40\%&57.26\%&$+\underline{53.90\%}\,\uparrow$\\
\midrule
\midrule
\multicolumn{10}{c}{\textbf{Qwen3-4B}} \\
\midrule
\textbf{Base} &Acc&0.9440&0.9462&0.9453&--&0.7890&0.958&0.9417&--\\
\midrule
\multirow{2}{*}{\textbf{SFT}} 
     &Acc   & 0.9400& 0.9386& 0.9390&$-\phantom{0}0.60\%\,\downarrow$&0.7850&0.9500&0.9400&$-\phantom{0}0.46\%\,\downarrow$\\
     &Compr.& 5.38\%& 8.33\%& -8.89\%&$+\phantom{0}1.61\%\,\uparrow$&-9.29\%&-10.42\%&-16.12\%&$-11.94\%\,\downarrow$\\
\midrule
\multirow{2}{*}{\textbf{DPO}} 
     &Acc   & 0.9260 & 0.9378 & 0.9492 &$-\phantom{0}0.75\%\,\downarrow$&0.7830&0.9510&0.9267&$-\phantom{0}0.93\%\,\downarrow$\\
     &Compr.& 37.16\%& 50.56\%& 47.86\%&$+\underline{45.22\%}\,\uparrow$ &79.01\%&76.37\%&77.49\%&$+\textbf{77.62\%}\,\uparrow$\\
\midrule
\multirow{2}{*}{\textbf{LC-R1}}
     &Acc   & 0.9280 & 0.9401 & 0.9438 &$-\phantom{0}0.79\%\,\downarrow$&0.7900&0.9520&0.9383&$-\phantom{0}0.28\%\,\downarrow$\\
     &Compr.& 40.43\%& 57.78\%& 61.67\%&$+\textbf{53.29\%}\,\uparrow$ &45.61\%&56.39\%&54.02\%&$+\underline{52.01\%}\,\uparrow$\\
\midrule
\multirow{2}{*}{\textbf{HAPO}} 
  &Acc   &0.9400&0.9469&0.9472&$-\phantom{0}\textbf{0.05\%}\,\downarrow$    &0.8060&0.9590&0.9417&$+\phantom{0}\textbf{0.60\%}\,\downarrow$\\
  &Compr.&9.74\%&3.77\%&-2.32\%    &$+\phantom{0}3.73\%\,\uparrow$&20.86\%&18.22\%&10.52\%&$+16.53\%\,\uparrow$\\
\midrule
\multirow{2}{*}{\textbf{ENTRA}} 
     &Acc   & 0.9320 & 0.9431 & 0.9453 &$-\phantom{0}\underline{0.50\%}\,\downarrow$&0.7930&0.9530&0.9483&$+\phantom{0}\underline{0.19\%}\,\uparrow$\\
     &Compr.& 38.27\%& 42.38\%& 39.44\%&$+40.03\%\,\uparrow$&53.93\%&49.87\%&46.46\%&$+50.09\%\,\uparrow$\\
\midrule
\bottomrule
\end{tabular}
}
\caption{Performance of accuracy and compression performance of different methods on qwen3 model. In the figure, Acc is the correct answer rate and Compr is the reduction ratio of their total reasoning tokens compared to the base model. AVG shows the average change (+ increase, - decrease) in accuracy and length compared to the base model. For the AVG column, the best performing score is marked in bold and the second best score is underlined.}
\label{tab:1}
\end{table*}

\section{Experiments}
\subsection{Training}
We apply our method to recent open-source reasoning models Qwen3-4B and Qwen3-8B. The training is conducted separately on both English and Chinese datasets, followed by evaluations on corresponding multilingual benchmarks. Details of the datasets are provided in Appendix A.
For fine-tuning, we adopt the LoRA method based on the Unsloth \cite{unsloth} codebase for parameter-efficient adaptation. We also include several representative baseline methods for comparison. More implementation details, including dataset splits and hyperparameters, are described in Appendix A.
\subsection{Evaluation}
For all models, we enable their reasoning capabilities to perform deep inference, strictly following the official recommendations and configurations. Specifically, we set the sampling parameters as follows: Temperature = 0.6, TopP = 0.95, TopK = 20, and the maximum generation length to 38,912 tokens. Additionally, we use the open-source evaluation framework EvalScope \cite{evalscope_2024}, and apply it uniformly across both English and Chinese datasets to ensure consistency.
\subsection{Results}
We compare ENTRA against several representative baselines under both Qwen3-4B and Qwen3-8B settings. The evaluation is conducted on three English datasets (MATH-500, GSM8K, MAWPS) and three Chinese datasets (Weekly12K, MATH23K, CMath), with accuracy (Acc) and compression ratio (Compr) as the primary metrics. Table~\ref{tab:1} presents the full results.

While SFT fits the shortest responses during training, its actual compression effect is limited, and in some cases even negative—showing that LoRA-based supervised training alone cannot effectively reduce redundant reasoning. DPO achieves high compression, especially on Chinese datasets (up to 77.62\%), but suffers from significant accuracy drops, highlighting its tendency to overfit short responses and overlook essential reasoning content.
LC-R1 offers strong compression on the 4B model, but its advantage is less consistent on the 8B model. HAPO generally preserves accuracy better, especially in Chinese settings, but achieves only modest compression.

ENTRA demonstrates a strong balance between accuracy and compression. On Qwen3-8B, it achieves a 37.88\% average reduction in length with the highest accuracy gains across datasets (+0.46\% English, +0.44\% Chinese). On Qwen3-4B, while slight accuracy drops are observed (-0.50\% English), compression remains competitive (40.03\%), and Chinese accuracy even improves. These results suggest that ENTRA not only reduces redundant reasoning, but also maintains or improves correctness, especially on larger models.

\begin{figure}[t!]
\centering
\includegraphics[width=1\columnwidth]{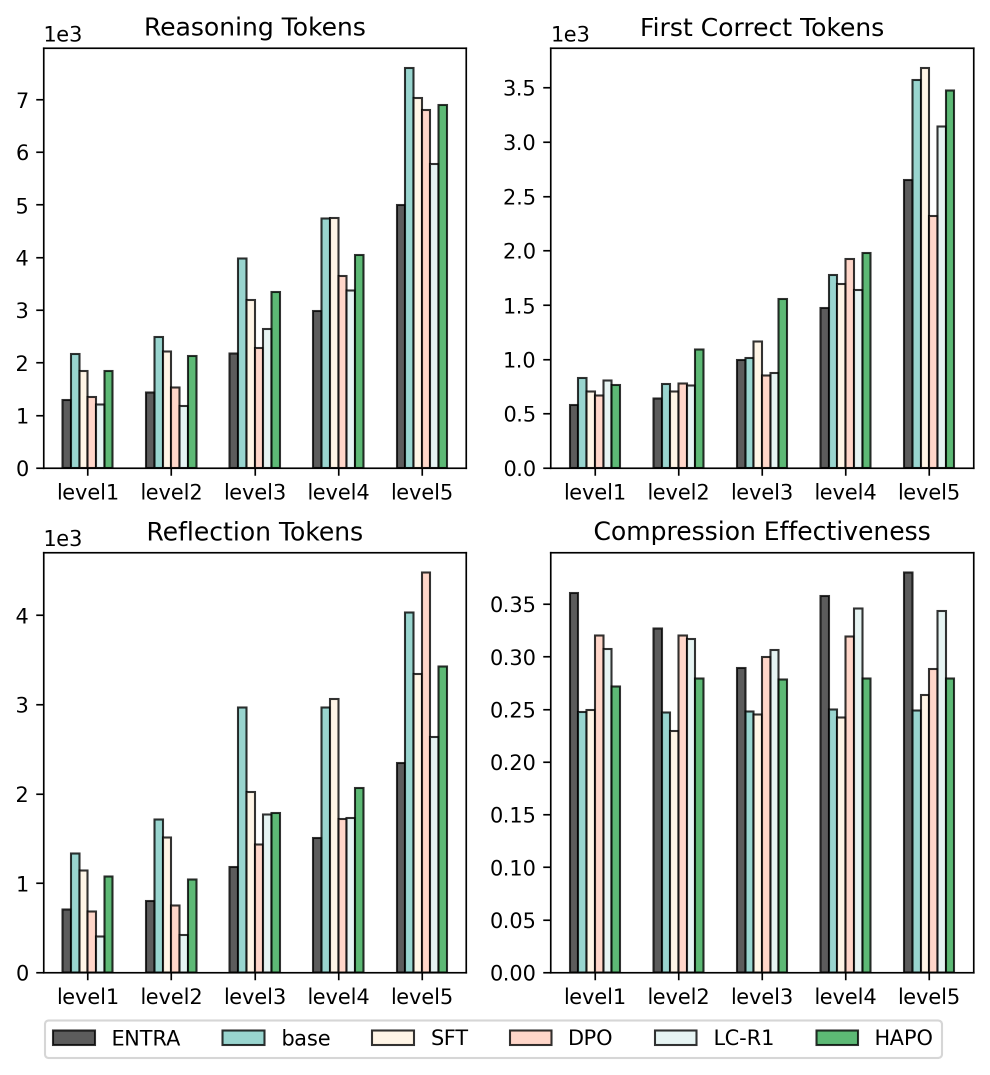} %
\caption{Compression analysis of each method based on Qwen3-8b}
\label{fig4}
\end{figure}
\begin{figure}[t!]
\centering
\includegraphics[width=1\columnwidth]{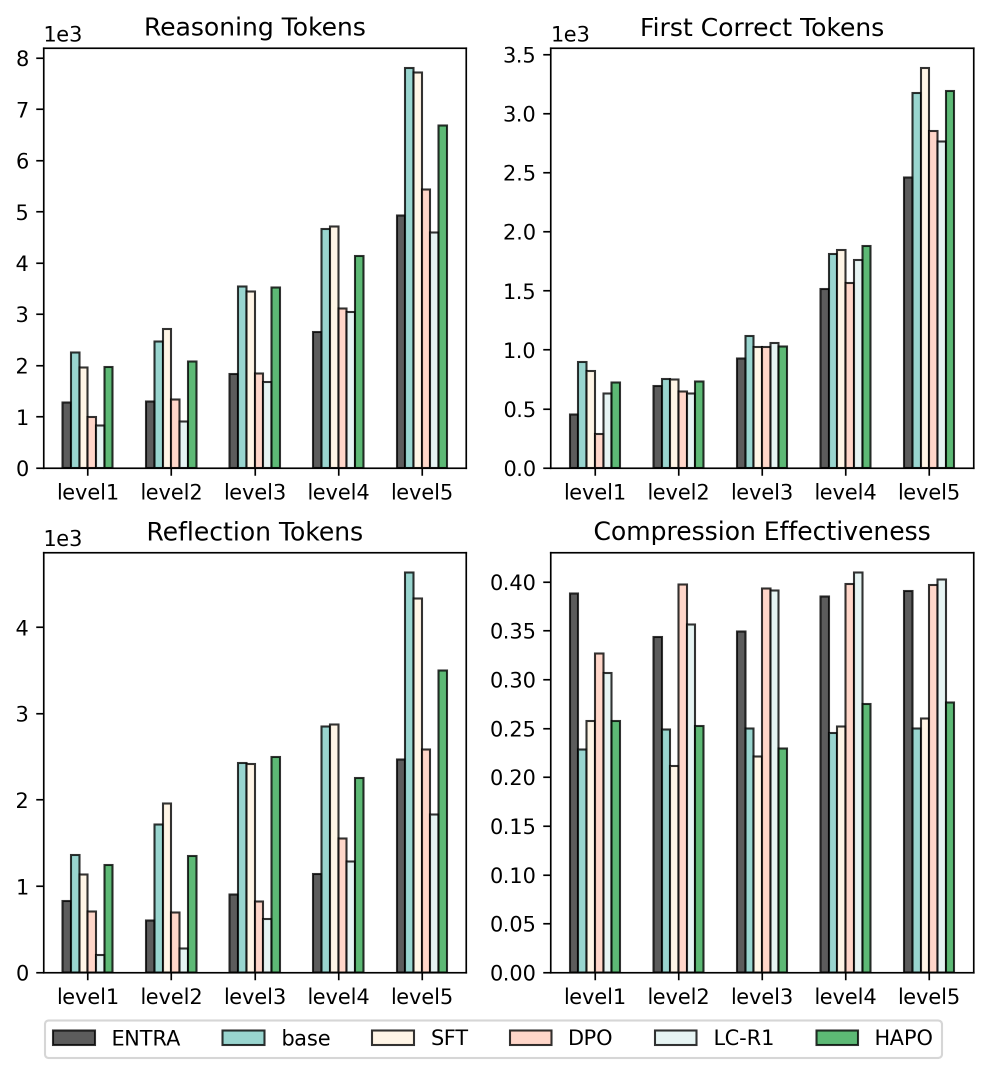} %
\caption{Compression analysis of each method based on Qwen3-4b}
\label{fig5}
\end{figure}

\subsection{Compression Analysis}
To more comprehensively evaluate the effectiveness of our method, we analyze its behavior from two key dimensions: reasoning accuracy and inference efficiency. In particular, we pay close attention to the model’s behavior before and after compression, and how redundant reasoning is reduced without compromising performance.

Following \citet{evalscope_2024} and \citet{chen2025think23overthinkingo1like}, we track three core metrics in the reasoning process: (1) the total number of tokens generated during reasoning (\textbf{Reasoning Tokens}, denoted $T$); (2) the index of the first token that reveals a correct answer (\textbf{First Correct Token}, denoted $\hat{T}$); and (3) the number of tokens that follow $\hat{T}$ until the end of generation, representing unnecessary elaboration, termed \textbf{Reflection Tokens} ($T - \hat{T}$).

\citet{chen2025think23overthinkingo1like} proposed using the ratio $\hat{T}/T$ to measure reasoning efficiency. However, this metric can be misleading: a model could delay producing the answer until near the end, and still score high despite inefficient reasoning. Therefore, we propose a new metric, \textbf{Compression Effectiveness} ($Ce$), which better captures the reduction of redundant tokens while preserving reasoning structure. It is defined as:
\begin{equation}
    Ce = \text{Compr} \cdot \frac{\hat{T}(T - \hat{T})}{T^2}
\end{equation}
where $\text{Compr}$ is the compression ratio relative to the base model. This formula jointly considers the position of the correct answer and the amount of redundant content that follows, balancing early correctness with compact generation. The metric reaches its maximum when the model answers correctly midway through reasoning and provides minimal yet meaningful elaboration afterward—closely aligning with desirable human-like inference behavior.

\begin{figure*}[t!]
\centering
\includegraphics[width=2\columnwidth]{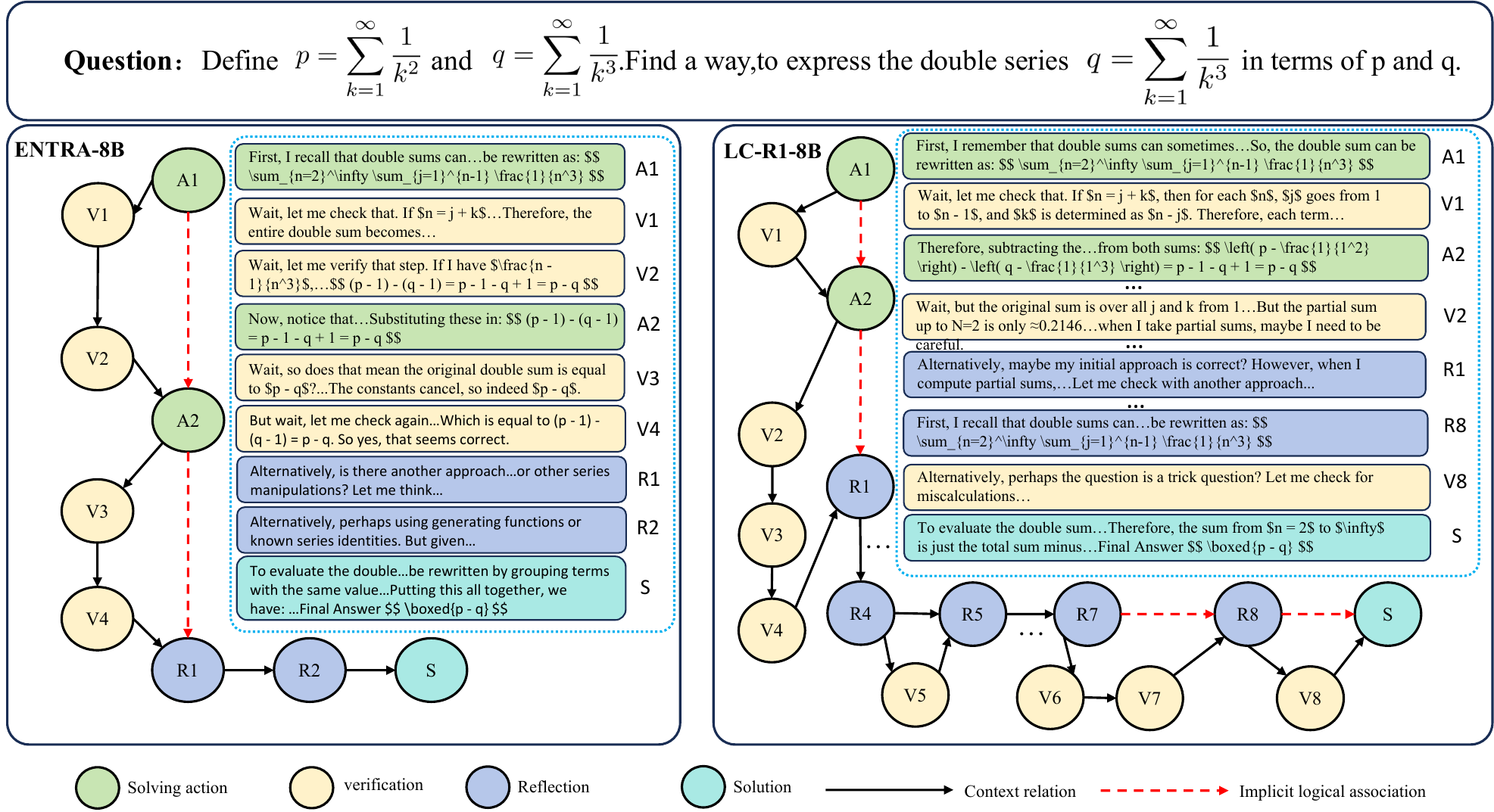} 
\caption{Case study comparing the reasoning trajectories of ENTRA-8B and LC-R1-8B. }
\label{case_study}
\end{figure*}

Figure~\ref{fig4} and Figure~\ref{fig5} visualize the performance of all methods on the \textbf{MATH-500} dataset across difficulty levels (Level 1–5), under the \textbf{Qwen3-8B} and \textbf{Qwen3-4B} backbones respectively.

We observe that \textbf{ENTRA} consistently maintains a shorter total reasoning length while achieving earlier correct answers and fewer reflection tokens, especially at higher difficulty levels. This indicates that ENTRA does not simply truncate outputs, but is capable of identifying and removing genuinely redundant inference steps. In comparison, \textbf{LC-R1} demonstrates strong compression ability on easier tasks but fails to maintain performance as problem difficulty increases, suggesting that its reduction is largely superficial and lacks structural pruning.

\textbf{DPO} and \textbf{SFT} show high variance. Though they sometimes achieve token reduction, this is often accompanied by degraded reasoning accuracy or coherence, implying that simple length constraints are insufficient to guide meaningful compression. \textbf{HAPO} performs well in maintaining accuracy, but its compression gains are modest and unstable across datasets and difficulty levels.

Overall, \textbf{ENTRA} achieves the highest Compression Effectiveness ($Ce$) in nearly all scenarios, demonstrating a better balance between brevity and reasoning depth. These results strongly support our hypothesis that entropy-guided redundancy control leads to more concise and interpretable reasoning paths without compromising correctness.

\subsection{Case Study}
We further compared the reasoning steps and the overall situation.Figure~\ref{case_study} conducts a comparative analysis between the currently highly performing LC-R1 method and the method proposed in this paper. The results show that, compared to the LC-R1 method, the method proposed in this paper can effectively maintain the accuracy of the final answer while significantly simplifying the redundant reflection and verification steps. This fully demonstrates the dual advantages of this method in terms of efficiency and reliability. More case studies are presented in Appendix C.

\section{Conclusion}
In this work, we propose \textbf{ENTRA}, a text-entropy-based training framework designed to reduce redundant reasoning in large language models. We systematically characterize repetitive generation as a form of redundancy, and define an optimization objective based on the entropy of low-importance tokens. To prevent over-pruning of essential reasoning steps, we introduce a lightweight bidirectional importance estimation (BIE) method to identify and exclude key tokens from redundancy computation.
By combining entropy-based reward with its theoretical upper bound, we derive a stable training signal for reinforcement learning using GRPO. Experimental results demonstrate that ENTRA significantly reduces unnecessary inference steps while maintaining, or even improving, model accuracy. Compared to heuristic length- or correctness-based supervision, our method more effectively guides the model to explore concise and efficient reasoning paths.
We believe this work offers a principled foundation for future research on redundancy-aware optimization in large language models, and paves the way toward more efficient and interpretable reasoning systems.
\section*{Limitations}
\paragraph{Unbiased estimation of bidirectional importance.}
The calculation of Bidirectional Importance Estimation adopts an approximate solution strategy. Although a hyperparameter $\lambda$ is introduced to optimize and adjust the interpretability of the model, this approximate calculation method still cannot completely eliminate bias, which may affect the accuracy of the bidirectional token importance estimation results to a certain extent. Future research will focus on exploring a more accurate bidirectional importance calculation framework and further improving the interpretability of the method to achieve more reliable quantitative estimation of the bidirectional importance of tokens.

\paragraph{Limitations of the Scope of Task Application.}
The core application scenario of the current research is limited to the reasoning compression task of mathematical reasoning problems. This limitation essentially stems from the structural characteristics of the bidirectional importance calculation formula—its design logic is more suitable for the semantic correlation and importance distribution characteristics of tokens in mathematical problems. Future research will expand the research boundary, explore the applicability of the bidirectional importance calculation method in the reasoning compression scenarios of more types of natural language processing tasks, and verify its cross-task transfer capability and generalization performance.

\bibliography{custom}

\appendix

\label{sec:appendix}

\section{Experiment details}
\subsection{Dataset}
We train and test our method on several mathematical datasets in both English and Chinese, including:\\
\textbf{English dataset.} GSM8K\citep{cobbe2021trainingverifierssolvemath}, which includes elementary school-level math problems.Math-500\cite{hendrycks2021measuringmathematicalproblemsolving} is a subset of 500 questions in the MATH benchmark test. MAWPS\cite{koncel-kedziorski-etal-2016-mawps}, which is a mathematics application question bank, is collected from multiple published datasets.\\
\textbf{Chinese dataset.} Weekly12K\cite{liang2022generalizingmathwordproblem} is a dataset of novel applied mathematical problems. Each problem in the dataset is annotated with a final value for the answer instead of the solution of the equation. MATH23K\cite{wang-etal-2017-deep} contains 23161 annotated expressions and answers for solutions of mathematical application problems. CMATH\cite{wei2023cmathlanguagemodelpass}, where is a dataset of Chinese primary school math word problems, these questions are from real Chinese workbooks and exams, which contains 1700 primary school math word problems.
\subsection{ENTRA Training}
We use our approach in the recent open source inference models Qwen3-4b and Qwen3-8b\cite{qwen3technicalreport}. And we trained our method on the English and Chinese datasets. In the training phase, for the English dataset, we use the first 1000 entries of the MATH dataset for training. For the Chinese dataset, we train on the first 1000 entries of the Weekly12K dataset. We fine-tune the model with lora based on the Unsloth codebase. By using Unsloth's memory optimization training, we are able to train GRPO with limited devices. In training, we set the lora rank to 128 and only train for 200 steps. For our BIAE module, we use qwen3-4b to calculate the token importance, set $\lambda=10^{-4}$, and set the importance threshold $\tau$to 20\%, that is, the top 20\% important tokens do not participate in the calculation of text information entropy.All experiments are conducted using 2 A800 80G GPUs.
\subsection{Prompt template for training and evaluation}
For all training experiment, we use the following prompt template:\\
\begin{prompt}[title={Prompt \thetcbcounter: Instruction}, label=prompt:Instruction]
Please reason step by step, and put your final answer within \text{\textbackslash boxed\{\}}.
\end{prompt}

\subsection{Baseline}
We discuss many works in efficient inference in Chapter 2, most of which currently focus on combined rewards for length and correctness. In addition to our method, we also introduce and implement a simple Baseline work to compare the superiority of our method. All methods use our dataset for training and evaluation results.\\
\textbf{SFT}.\quad In this baseline, we use vLLM to sample 8 outputs for each example and select the shortest one for supervised fine-tuning. The sampling parameters are: temperature = 0.6, top\_p = 0.95, top\_k = 20, max\_tokens = 32,768. During training, we set lora\_rank = 128, maximum training sequence length = 8,192, learning rate = 2e-4, and number of training epochs = 5.\\
\textbf{DPO}\cite{rafailov2024directpreferenceoptimizationlanguage}.\quad In this baseline, we use vLLM to sample 8 outputs for each example, selecting the shortest sample as the “chosen” output and the longest sample as the “rejected” output. The sampling parameters are: temperature = 0.6, top\_p = 0.95, top\_k = 20, and max\_tokens = 32,768. During training, we set lora\_rank = 128, the maximum training sequence length to 4,096, the learning rate to 2e-4, and the number of training epochs = 5.\\
\textbf{LC-R1}.\cite{cheng2025optimizinglengthcompressionlarge}\quad A method that combines length reward with compression reward. We apply this method to train and evaluate on our dataset. Following the author’s approach, we perform full-parameter training with a maximum training sequence length of 4096, a learning rate of 2e-6, num\_generations = 4, and train for 1 epoch.\\
\textbf{HAPO}.\cite{huang2025hapotraininglanguagemodels}\quad A method that combines length reward and correctness reward based on historical states. Similarly, we perform full-parameter training with a maximum training sequence length of 4096, a weight learning rate (w\_lr) of 1.0, a cosine learning rate schedule (type\_lr = cosine), num\_generations = 2, and train for 1 epoch.

\begin{figure*}[t!]
\centering
\includegraphics[width=2\columnwidth]{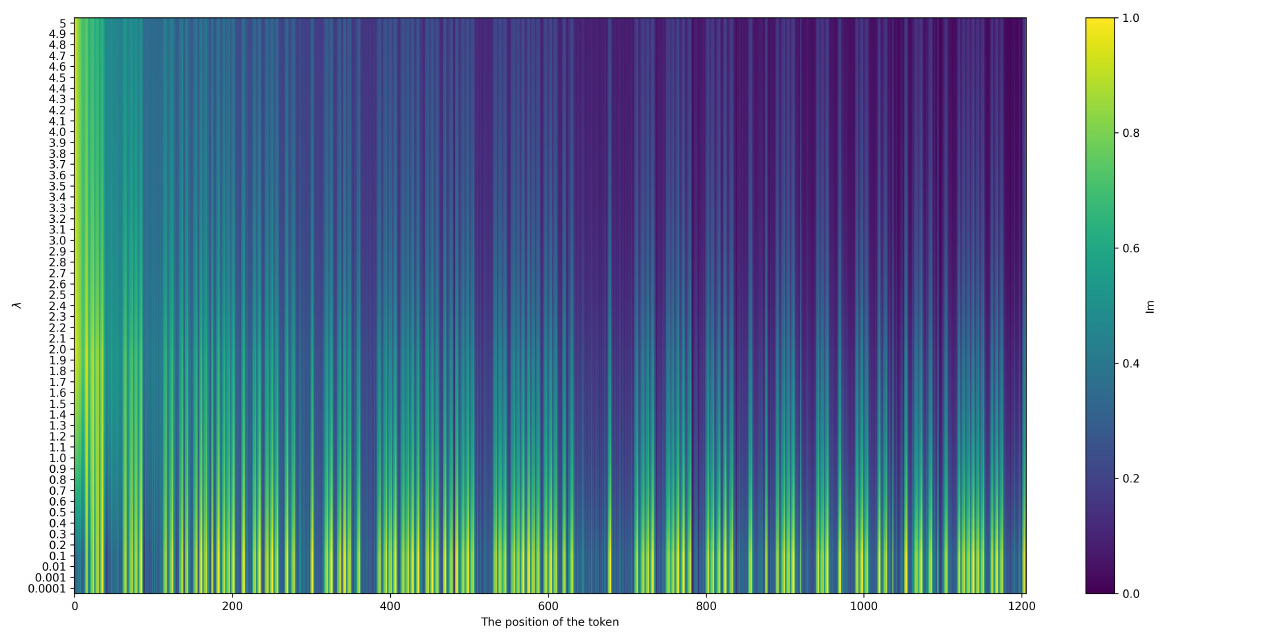} 
\caption{Different $\lambda$, token importance allocation results}
\label{Figure6}
\end{figure*}

\section{Discussion on the value of $\lambda$}
In this section, we focus on the selection of the hyperparameter $\lambda$ in the BIE module. It should be noted that due to the adoption of an approximation method in the calculation of future contribution values in this paper, the estimation of the importance score $\log P(x_i \mid x_{<i}, x_{>i})$ is biased and cannot satisfy the unbiasedness assumption. To verify the interpretability of the proposed setup and determine the optimal value of $\lambda$, we visualize the model's importance assignment results for each input token when $\lambda$ decays from 5 to $10^{-4}$, as shown in Figure 6. 

The experimental results in Figure 6 demonstrate that there is an obvious correlation between the value of $\lambda$ and the assignment of token importance: as the value of $\lambda$ increases, the model’s importance score for tokens in the early part of the sequence increases significantly, while the importance score for tokens in the later part decreases accordingly. In the task setup of this paper, tokens in the early part of the sequence correspond to the question stem of mathematical problems. From a semantic perspective, question stem tokens are key components that guide LRMs to understand the core of the task, and their importance should be fully captured. 

Based on the above analysis, we finally set $\lambda = 2$ in the subsequent experiments of this paper. This value can effectively enhance the importance of key tokens in the question stem while retaining the information weight of core tokens in the subsequent sequence, thus achieving accurate capture of key information at different positions in mathematical problems.

\section{Case Study}
We conduct a case study to compare the full reasoning processes of ENTRA, LC-R1, and the original Qwen3-8B model, in order to more thoroughly illustrate the differences presented in Figure 5. Inspired by the approach in \cite{luo2025deconstructinglongchainofthoughtstructured}, we similarly decompose the reasoning chain of LRM into three types of actions:
\begin{itemize}
    \item Actions that directly carry out the problem-solving process(Solving action).
    \item Actions that verify the correctness of a specific step(verification), typically beginning with “Wait”, indicating a self-check on the validity of a previous step;
    \item Actions that reflect on the feasibility of alternative solutions(Reflection), usually starting with “Alternatively”, intended to explore other potential solution paths.
\end{itemize}

As shown in \ref{case:1}, \ref{case:2}, and \ref{case:3}, we present the complete reasoning trajectories of the three methods. It is evident that both LC-R1 and the original Qwen3-8B frequently engage in repetitive exploratory steps at certain points in the reasoning process. These steps are often unnecessary and redundant. Although they eventually produce correct answers, the excessive reasoning leads to decreased readability and significant computational overhead.
In contrast, the ENTRA method significantly reduces superfluous verification and reflection actions, while still maintaining correctness in the final answers. This demonstrates ENTRA’s superior reasoning efficiency and clarity.

\begin{CASE}[label=case:1,title=Case \thetcbcounter: \\The Case of ENTRA for the following problem:\\Define  {$p = \sum_{k = 1}^\infty \frac{1}{k^2} \quad \text{and} \quad q = \sum_{k = 1}^\infty \frac{1}{k^3}.$}
Find a way to write{$\sum_{j = 1}^\infty \sum_{k = 1}^\infty \frac{1}{(j + k)^3}$}in terms of $p$ and $q$.]

\textless{}think\textgreater{}

Okay, so I need to find a way to express the double sum $\sum_{j = 1}^\infty \sum_{k = 1}^\infty \frac{1}{(j + k)^3}$ in terms of $p$ and $q$, where $p = \sum_{k = 1}^\infty \frac{1}{k^2}$ and $q = \sum_{k = 1}^\infty \frac{1}{k^3}$. Let me start by understanding the problem and thinking about possible strategies.

First, I recall that double sums can sometimes be simplified by changing variables or recognizing patterns. The given double sum involves terms of the form $\frac{1}{(j + k)^3}$, so maybe I can make a substitution to simplify this expression. Let me consider letting $n = j + k$. If I fix $n$, then for each $n$, the possible values of $j$ and $k$ are such that $j$ ranges from 1 to $n - 1$ and $k = n - j$. However, since both $j$ and $k$ start at 1, $n$ must be at least 2. So, the double sum can be rewritten as:

$$
\sum_{n=2}^\infty \sum_{j=1}^{n-1} \frac{1}{n^3}
$$

Wait, let me check that. If $n = j + k$, then for each $n$, the number of pairs $(j, k)$ such that $j + k = n$ is $n - 1$, since $j$ can be 1 to $n - 1$ and $k$ would be $n - j$. Therefore, the inner sum over $j$ for each fixed $n$ is just $(n - 1)$ terms each equal to $\frac{1}{n^3}$. Therefore, the entire double sum becomes:

$$
\sum_{n=2}^\infty \frac{n - 1}{n^3} = \sum_{n=2}^\infty \left( \frac{1}{n^2} - \frac{1}{n^3} \right)
$$

Wait, let me verify that step. If I have $\frac{n - 1}{n^3}$, that can be split as $\frac{n}{n^3} - \frac{1}{n^3} = \frac{1}{n^2} - \frac{1}{n^3}$. Yes, that seems correct. Therefore, the original double sum simplifies to:

$$
\sum_{n=2}^\infty \left( \frac{1}{n^2} - \frac{1}{n^3} \right) = \left( \sum_{n=2}^\infty \frac{1}{n^2} \right) - \left( \sum_{n=2}^\infty \frac{1}{n^3} \right)
$$

Now, notice that $p = \sum_{n=1}^\infty \frac{1}{n^2}$ and $q = \sum_{n=1}^\infty \frac{1}{n^3}$. Therefore, the sums from $n=2$ to infinity are just $p - 1$ and $q - 1$ respectively. Substituting these in:

$$
(p - 1) - (q - 1) = p - 1 - q + 1 = p - q
$$
Wait, so does that mean the original double sum is equal to $p - q$? Let me check my steps again to make sure I didn't make a mistake.

Starting with the double sum over $j$ and $k$ from 1 to infinity of $1/(j + k)^3$. Changing variables to $n = j + k$, which gives for each $n \geq 2$, the number of pairs is $n - 1$, so the inner sum over $j$ becomes $(n - 1)/n^3$. Then splitting into $1/n^2 - 1/n^3$, summing over $n \geq 2$, which gives $p - 1 - (q - 1)$. The constants cancel, so indeed $p - q$. 

But let me test with some small numbers. Let me compute partial sums to see if this holds.

For example, take the first few terms of the original double sum. Let me compute the sum up to j=1, k=1:

When j=1, k=1: $1/(1+1)^3 = 1/8$.

j=1, k=2: $1/3^3 = 1/27$.

j=1, k=3: $1/4^3 = 1/64$.

Similarly for j=2, k=1: same as j=1, k=2: 1/27.

So the total for n=2: 2*(1/8) = 1/4? Wait, no. Wait when n=2, j=1, k=1. So only one term, which is 1/8. Wait, n=2: j=1, k=1. n=3: j=1,k=2 and j=2,k=1. So two terms, each 1/27. So total for n=3 is 2/27. Similarly n=4: 3 terms each 1/64, so 3/64. Then the total sum up to n=4 would be 1/8 + 2/27 + 3/64 + ... 

But according to my formula, the total sum would be p - q. Let me compute p and q approximately.

p is $\pi^2/6 \approx 1.6449$, and q is $\zeta(3) \approx 1.202$. So $p - q \approx 0.4429$.

Now, let me compute the partial sum up to n=4:

$1/8 + 2/27 + 3/64 \approx 0.125 + 0.07407 + 0.046875 \approx 0.2459$. Then adding more terms would approach 0.4429. So seems plausible. Let me check for n=5, which would add 4/125 = 0.032, so total becomes ~0.2779. Then n=6: $5/216 \approx 0.0231$, total ~0.301. n=7: $6/343 \approx 0.0175$, total ~0.318. Continuing, n=8: $7/512 \approx 0.0137$, total ~0.331. n=9: $8/729 \approx 0.01097$, total ~0.342. n=10: $9/1000=0.009$, total ~0.351. So seems like it's approaching 0.44... So maybe the formula is correct.

Therefore, my conclusion is that the original double sum is equal to $p - q$. Therefore, the answer is $\boxed{p - q}$.

But wait, let me check again. Because in my derivation, the double sum becomes $\sum_{n=2}^\infty (1/n^2 - 1/n^3)$. Which is equal to (p - 1) - (q - 1) = p - q. So yes, that seems correct.

Alternatively, is there another approach? Maybe using integral representations or other series manipulations? Let me think.

Alternatively, perhaps using generating functions or known series identities. But given that my first approach seems straightforward and the partial sums seem to check out, I think that this is the correct answer. Therefore, the answer is $p - q$.

**Final Answer**
\boxed{p - q}

\textless{}/think\textgreater{}

To evaluate the double sum

$$
\sum_{j = 1}^\infty \sum_{k = 1}^\infty \frac{1}{(j + k)^3},
$$

we begin by observing that the expression depends only on the sum $ n = j + k $. For a fixed $ n \geq 2 $, the number of integer pairs $(j, k)$ such that $ j + k = n $ is $ n - 1 $, since $ j $ ranges from 1 to $ n - 1 $, and $ k = n - j $.

Thus, the original double sum can be rewritten by grouping terms with the same value of $ n $:

$$
\sum_{j = 1}^\infty \sum_{k = 1}^\infty \frac{1}{(j + k)^3} = \sum_{n = 2}^\infty \sum_{j = 1}^{n - 1} \frac{1}{n^3}.
$$

Since the inner sum has $ n - 1 $ terms, each of which is $ \frac{1}{n^3} $, this simplifies to:

$$
\sum_{n = 2}^\infty \frac{n - 1}{n^3} = \sum_{n = 2}^\infty \left( \frac{1}{n^2} - \frac{1}{n^3} \right).
$$

Now, we recognize that the sum $ \sum_{n=2}^\infty \frac{1}{n^2} $ is the tail of the well-known Riemann zeta function $ p = \sum_{n=1}^\infty \frac{1}{n^2} $, so:

$$
\sum_{n=2}^\infty \frac{1}{n^2} = p - 1.
$$

Similarly, $ \sum_{n=2}^\infty \frac{1}{n^3} = q - 1 $, where $ q = \sum_{n=1}^\infty \frac{1}{n^3} $.

Putting this all together, we have:

$$
\sum_{j = 1}^\infty \sum_{k = 1}^\infty \frac{1}{(j + k)^3} = (p - 1) - (q - 1) = p - q.
$$

---\textbf{Final Answer}
$$
\boxed{p - q}
$$
\end{CASE}

\begin{CASE}[label=case:2,title=Case \thetcbcounter: \\The Case of LC-R1 for the following problem:\\Define  {$p = \sum_{k = 1}^\infty \frac{1}{k^2} \quad \text{and} \quad q = \sum_{k = 1}^\infty \frac{1}{k^3}.$}
Find a way to write{$\sum_{j = 1}^\infty \sum_{k = 1}^\infty \frac{1}{(j + k)^3}$}in terms of $p$ and $q$.]

\textless{}think\textgreater{}

Okay, so I need to find a way to express the double sum $\sum_{j = 1}^\infty \sum_{k = 1}^\infty \frac{1}{(j + k)^3}$ in terms of $p$ and $q$, where $p = \sum_{k=1}^\infty \frac{1}{k^2}$ and $q = \sum_{k=1}^\infty \frac{1}{k^3}$. Hmm, let me think about how to approach this.

First, I remember that double sums can sometimes be simplified by changing variables or by recognizing patterns. The term $(j + k)$ suggests that maybe I can make a substitution to combine the indices. Let me try to think about how to reindex this sum.

Let me denote $n = j + k$. Then for each fixed $n$, $j$ can range from 1 to $n - 1$, and $k = n - j$. However, since both $j$ and $k$ start from 1, $n$ would range from 2 to infinity. So, the double sum can be rewritten as:

$$
\sum_{n=2}^\infty \sum_{j=1}^{n-1} \frac{1}{n^3}
$$

Wait, let me check that. If $n = j + k$, then for each $n$, $j$ goes from 1 to $n - 1$, and $k$ is determined as $n - j$. Therefore, each term $\frac{1}{(j + k)^3}$ becomes $\frac{1}{n^3}$, and the inner sum is over all $j$ from 1 to $n - 1$. Therefore, the double sum becomes:

$$
\sum_{n=2}^\infty \left( \frac{1}{n^3} \sum_{j=1}^{n - 1} 1 \right) = \sum_{n=2}^\infty \frac{n - 1}{n^3}
$$

Simplifying that, it's:

$$
\sum_{n=2}^\infty \left( \frac{1}{n^2} - \frac{1}{n^3} \right) = \sum_{n=2}^\infty \frac{1}{n^2} - \sum_{n=2}^\infty \frac{1}{n^3}
$$

But we know that $p = \sum_{n=1}^\infty \frac{1}{n^2}$ and $q = \sum_{n=1}^\infty \frac{1}{n^3}$. Therefore, subtracting the first term (which is $n=1$) from both sums:

$$
\left( p - \frac{1}{1^2} \right) - \left( q - \frac{1}{1^3} \right) = p - 1 - q + 1 = p - q
$$

Wait, that seems too straightforward. Let me check again. If I have $\sum_{n=2}^\infty \frac{1}{n^2} = p - 1$ and $\sum_{n=2}^\infty \frac{1}{n^3} = q - 1$, so the difference is $(p - 1) - (q - 1) = p - q$. So the original double sum is equal to $p - q$? Hmm, but that seems surprisingly simple. Let me verify this with an example or check for possible mistakes.

Wait, let me think again. Let me take a small case. Let's compute the double sum up to some small N and see if it matches p - q. Let's say N=2. Then the original double sum would be for j and k from 1 to 2. Wait, but the original sum is over all j and k from 1 to infinity. If I take N=2, then the terms would be:

For j=1, k=1: $1/(1+1)^3 = 1/8$

j=1, k=2: $1/(1+2)^3 = 1/27$

j=2, k=1: $1/(2+1)^3 = 1/27$

j=2, k=2: $1/(2+2)^3 = 1/64$

So total sum up to N=2 is $1/8 + 1/27 + 1/27 + 1/64 \approx 0.125 + 0.037 + 0.037 + 0.0156 \approx 0.2146.$

Now, p is $\pi^2/6 \approx 1.6449$, q is $\zeta(3) \approx 1.202$. Therefore, $p - q \approx 0.4429$. But the partial sum up to N=2 is only $\approx0.2146$. So if we take more terms, maybe it approaches p - q? Let me check N=3.

For N=3, the original double sum would include all terms where j and k go up to 3. But actually, since the original sum is over all j and k, but when I reindexed earlier, I considered n from 2 to infinity. However, when I take partial sums, maybe I need to be careful. Alternatively, maybe my initial approach is correct?

Wait, but according to my previous calculation, the entire double sum equals p - q. However, when I compute partial sums, they seem to be smaller. But maybe as N approaches infinity, the partial sums approach p - q? Let me check with another approach.

Alternatively, perhaps I made a mistake in the reindexing. Let me go back.

Original double sum:

$$
\sum_{j=1}^\infty \sum_{k=1}^\infty \frac{1}{(j + k)^3}
$$

Changing variables to n = j + k and m = j (for example). Then for each $n \geq 2$, j can be from 1 to n - 1, so k = n - j. Therefore, the double sum becomes:

$$
\sum_{n=2}^\infty \sum_{j=1}^{n - 1} \frac{1}{n^3} = \sum_{n=2}^\infty \frac{n - 1}{n^3} = \sum_{n=2}^\infty \left( \frac{1}{n^2} - \frac{1}{n^3} \right)
$$

Which is equal to:

$$
\left( \sum_{n=2}^\infty \frac{1}{n^2} \right) - \left( \sum_{n=2}^\infty \frac{1}{n^3} \right) = \left( p - 1 \right) - \left( q - 1 \right) = p - q
$$

So according to this, the answer is p - q. But why does the partial sum up to N=2 give only 0.2146 while p - q is approximately 0.4429? Because the partial sums are only up to n=2, but the full sum includes all n from 2 to infinity. Let me compute the partial sum up to n=3.

For n=2: $sum_{j=1}^1 1/2^3 = 1/8 = 0.125$

For n=3: $sum_{j=1}^2 1/3^3 = 2/27 \approx 0.07407$

For n=4: $sum_{j=1}^3 1/4^3 = 3/64 \approx 0.046875$

For n=5: $sum_{j=1}^4 1/5^3 = 4/125 = 0.032$

Adding these up:

n=2: 0.125

n=3: $0.125 + 0.07407 \approx 0.19907$

n=4: $+ 0.046875 \approx 0.2459$

n=5: $+ 0.032 \approx 0.2779$

Continuing:

n=6: $5/216 \approx 0.0231, total \approx 0.301$

n=7: $6/343 \approx 0.0175, total \approx 0.3185$

n=8: $7/512 \approx 0.01367, total \approx 0.332$

n=9: $8/729 \approx 0.01097, total \approx 0.343$

n=10: $9/1000 = 0.009, total \approx 0.352$

Continuing this, the partial sums approach $p - q \approx 0.4429$. So it seems plausible. Therefore, maybe my initial calculation is correct? Then why does the problem say "Find a way to write... in terms of p and q"? Maybe it's just p - q?

But that seems too straightforward. Let me check with another approach to confirm.

Alternatively, maybe there's a mistake in my reindexing? Let me think again.

Original sum is over all j and $k \geq 1$. So for each j and k, we have (j + k). If I fix n = j + k, then for each $n \geq 2$, the number of pairs (j, k) with j + k = n is n - 1. Therefore, the double sum is equal to $sum_{n=2}^\infty (n - 1)/n^3$. Which is $sum_{n=2}^\infty (1/n^2 - 1/n^3) = (sum_{n=1}^\infty 1/n^2 - 1) - (sum_{n=1}^\infty 1/n^3 - 1) ) = p - 1 - q + 1 = p - q$. So that's correct.

Therefore, the answer is p - q? But let me check if there's another way to think about this problem, maybe using integrals or other series manipulations, to confirm.

Alternatively, perhaps I can relate the double sum to p and q through some other method. For example, consider that:

$$
\sum_{j=1}^\infty \sum_{k=1}^\infty \frac{1}{(j + k)^3} = \sum_{j=1}^\infty \sum_{k=1}^\infty \frac{1}{(j + k)^3}
$$

Maybe I can use the integral representation of $1/n^3$? For example, since $\frac{1}{n^3} = \int_0^1 \int_0^1 \int_0^1 x^{n-1} y^{n-1} z^{n-1} dx dy dz$? Wait, maybe not. Alternatively, recall that $\frac{1}{n^3} = \int_0^1 \int_0^1 \int_0^1 x^{n-1} dx dy dz$? Wait, perhaps not straightforward.

Alternatively, recall that $1/(j + k)^3$ can be expressed as an integral. For example, we know that $\frac{1}{n^3} = \int_0^1 \int_0^1 \int_0^1 x^{n-1} dx dy dz$? Wait, actually, integrating $x^{n-1}$ from 0 to 1 gives 1/n. So integrating three times? Maybe:

$$
\frac{1}{n^3} = \int_0^1 \int_0^1 \int_0^1 x^{n-1} y^{n-1} z^{n-1} dx dy dz
$$

But I'm not sure if that helps here. Alternatively, perhaps using generating functions?

Alternatively, maybe consider that the original sum can be related to the square of some series? But since the terms are $1/(j + k)^3$, not 1/(j k) or something else. Maybe not directly.

Alternatively, think about symmetry. The original sum is symmetric in j and k, so maybe we can write it as 2 times the sum over j < k plus the sum over j = k? But since j and k start at 1, the diagonal terms where j = k would be when j = k, so n = 2j. So:

$$
\sum_{j=1}^\infty \sum_{k=1}^\infty \frac{1}{(j + k)^3} = \sum_{j=1}^\infty \frac{1}{(2j)^3} + 2 \sum_{j=1}^\infty \sum_{k=1}^{j - 1} \frac{1}{(j + k)^3}
$$

But I don't know if this helps. Let me compute the diagonal terms: $sum_{j=1}^\infty 1/(2j)^3 = (1/8) sum_{j=1}^\infty 1/j^3 = q/8$. Then the other part is twice the sum over $j > k of 1/(j + k)^3$. But I don't see how this connects to p and q directly. Maybe this approach isn't helpful.

Alternatively, going back to my initial approach, which seems to give p - q. But maybe there's a mistake in assuming that the double sum equals $sum_{n=2}^\infty (n - 1)/n^3$. Let me check for n=2: when n=2, there is only one pair (j=1, k=1), so the term is 1/8. According to$ (n - 1)/n^3$, that is 1/8, which matches. For n=3, there are two pairs: (1,2) and (2,1), so 2*(1/27) = 2/27, which matches $(3 - 1)/3^3 = 2/27$. So that seems correct. Therefore, the double sum is indeed $sum_{n=2}^\infty (n - 1)/n^3 = p - q$. Therefore, the answer is p - q? But the problem says "Find a way to write... in terms of p and q." So maybe that's the answer? But let me check some references or think if there's another way.

Wait, but I recall that sometimes double sums can be tricky because of convergence issues, but in this case, since all terms are positive, and we've reindexed properly, it should be okay. Therefore, maybe the answer is indeed p - q. But I need to make sure that I didn't overlook something.

Wait, let me check with another perspective. Let me consider that:

$$
\sum_{j=1}^\infty \sum_{k=1}^\infty \frac{1}{(j + k)^3} = \sum_{j=1}^\infty \sum_{k=1}^\infty \frac{1}{(j + k)^3}
$$

Let me make substitution m = j + k. For each $m \geq 2$, the number of pairs (j, k) with j + k = m is m - 1. Therefore, the sum becomes:

\[
\begin{aligned}
\sum_{m=2}^\infty \frac{m - 1}{m^3} 
&= \sum_{m=2}^\infty \left( \frac{1}{m^2} - \frac{1}{m^3} \right) \\
&= \left( \sum_{m=1}^\infty \frac{1}{m^2} - 1 \right) - \left( \sum_{m=1}^\infty \frac{1}{m^3} - 1 \right) \\
&= p - 1 - q + 1 \\
&= p - q
\end{aligned}
\]

Yes, same result. So unless there's a mistake in my reasoning, the answer is p - q. However, I need to recall if there are any standard results or if this is a known identity. Wait, but maybe the original problem is from a source where the answer is different? Or perhaps I made a miscalculation.

Alternatively, perhaps the question is a trick question? Let me check for miscalculations.

Wait, let me check the step where I say that the number of pairs (j, k) with j + k = m is m - 1. For $m \geq 2$, yes. For example, m=2: (1,1). m=3: (1,2), (2,1). m=4: (1,3), (2,2), (3,1). So indeed m - 1 terms. Therefore, the count is correct.

Therefore, the conclusion is that the double sum equals p - q. Hence, the answer is p - q. Therefore, boxed, it would be \boxed{p - q}.

But wait, I'm a bit uncertain because sometimes when dealing with double sums, especially with different indices, there might be some hidden steps or considerations. However, given that multiple approaches lead to the same conclusion, and the partial sums seem to approach p - q as n increases, I think this is correct. Therefore, I think the answer is p - q.

**Final Answer**
\boxed{p - q}

\textless{}/think\textgreater{}

To evaluate the double sum  
$$
\sum_{j = 1}^\infty \sum_{k = 1}^\infty \frac{1}{(j + k)^3},
$$  
we begin by observing the structure of the expression. The term $(j + k)$ suggests that a change of variables might simplify the sum. Let us define a new variable:

$$
n = j + k.
$$

For fixed $n \geq 2$, the values of $j$ and $k$ satisfying $j + k = n$ are such that $j$ ranges from $1$ to $n - 1$, and $k = n - j$. Thus, for each $n$, there are exactly $n - 1$ such pairs $(j, k)$. Therefore, the double sum can be reindexed as:

$$
\sum_{j = 1}^\infty \sum_{k = 1}^\infty \frac{1}{(j + k)^3} = \sum_{n = 2}^\infty \sum_{j = 1}^{n - 1} \frac{1}{n^3}.
$$

Since the inner sum is simply over $n - 1$ terms each equal to $\frac{1}{n^3}$, we can simplify:

$$
\sum_{n = 2}^\infty \frac{n - 1}{n^3} = \sum_{n = 2}^\infty \left( \frac{1}{n^2} - \frac{1}{n^3} \right).
$$

Now, we can split this into two separate sums:

$$
\sum_{n = 2}^\infty \frac{1}{n^2} - \sum_{n = 2}^\infty \frac{1}{n^3}.
$$

Recall the definitions of $p$ and $q$:

$$
p = \sum_{n = 1}^\infty \frac{1}{n^2}, \quad q = \sum_{n = 1}^\infty \frac{1}{n^3}.
$$

Therefore, the sum from $n = 2$ to $\infty$ is just the total sum minus the first term:

$$
\sum_{n = 2}^\infty \frac{1}{n^2} = p - \frac{1}{1^2} = p - 1, \quad \sum_{n = 2}^\infty \frac{1}{n^3} = q - \frac{1}{1^3} = q - 1.
$$

Substituting back:

$$
\sum_{j = 1}^\infty \sum_{k = 1}^\infty \frac{1}{(j + k)^3} = (p - 1) - (q - 1) = p - q.
$$

---\textbf{Final Answer}

$$
\boxed{p - q}
$$
\end{CASE}

\begin{CASE}[label=case:3,title=Case \thetcbcounter: \\The Case of Qwen3-8B for the following problem:\\Define  {$p = \sum_{k = 1}^\infty \frac{1}{k^2} \quad \text{and} \quad q = \sum_{k = 1}^\infty \frac{1}{k^3}.$}
Find a way to write{$\sum_{j = 1}^\infty \sum_{k = 1}^\infty \frac{1}{(j + k)^3}$}in terms of $p$ and $q$.]

\textless{}think\textgreater{}

Okay, so I need to find a way to express the double sum $\sum_{j = 1}^\infty \sum_{k = 1}^\infty \frac{1}{(j + k)^3}$ in terms of $p$ and $q$, where $p = \sum_{k = 1}^\infty \frac{1}{k^2}$ and $q = \sum_{k = 1}^\infty \frac{1}{k^3}$. Hmm, let me think about how to approach this.

First, I remember that double sums can sometimes be simplified by changing variables or by recognizing patterns. The given double sum is over $j$ and $k$ starting from 1, and the term is $\frac{1}{(j + k)^3}$. Maybe I can make a substitution to combine the indices $j$ and $k$ into a single variable. Let me try that.

Let me denote $n = j + k$. Since both $j$ and $k$ start at 1, the smallest value of $n$ is $1 + 1 = 2$. Then, for each $n$, the number of pairs $(j, k)$ such that $j + k = n$ is $n - 1$. For example, when $n = 2$, there's only one pair: (1,1). When $n = 3$, there are two pairs: (1,2) and (2,1). So in general, for each $n \geq 2$, there are $n - 1$ terms in the double sum where $j + k = n$.

Therefore, the original double sum can be rewritten as:

$$
\sum_{n=2}^\infty \frac{n - 1}{n^3}
$$

Wait, is that right? Let me check. If I fix $n = j + k$, then each term $\frac{1}{n^3}$ is multiplied by the number of ways to write $n$ as $j + k$ with $j, k \geq 1$. Which is indeed $n - 1$. So yes, that seems correct. Therefore, the double sum simplifies to:

$$
\sum_{n=2}^\infty \frac{n - 1}{n^3} = \sum_{n=2}^\infty \left( \frac{n}{n^3} - \frac{1}{n^3} \right ) = \sum_{n=2}^\infty \left( \frac{1}{n^2} - \frac{1}{n^3} \right )
$$

So that simplifies to:

$$
\sum_{n=2}^\infty \frac{1}{n^2} - \sum_{n=2}^\infty \frac{1}{n^3}
$$

But $p = \sum_{n=1}^\infty \frac{1}{n^2}$, so $\sum_{n=2}^\infty \frac{1}{n^2} = p - 1$. Similarly, $q = \sum_{n=1}^\infty \frac{1}{n^3}$, so $\sum_{n=2}^\infty \frac{1}{n^3} = q - 1$. Therefore, substituting back, we have:

$$
(p - 1) - (q - 1) = p - 1 - q + 1 = p - q
$$

Wait, so does that mean the original double sum is equal to $p - q$? But let me check again because that seems too straightforward. Let me verify with some numbers.

Alternatively, maybe I made a mistake in the substitution. Let me re-examine the steps.

Original double sum: $\sum_{j=1}^\infty \sum_{k=1}^\infty \frac{1}{(j + k)^3}$.

Changing variables to $n = j + k$, but for each $n$, the number of pairs is $n - 1$. Therefore, the sum becomes $\sum_{n=2}^\infty \frac{n - 1}{n^3}$. Then splitting into two sums:

$$
\sum_{n=2}^\infty \frac{1}{n^2} - \sum_{n=2}^\infty \frac{1}{n^3} = (p - 1) - (q - 1) = p - q
$$

Yes, that seems correct. But let me check with specific values. Let me compute partial sums numerically to see if it aligns.

First, let me recall that $p = \frac{\pi^2}{6} \approx 1.6449$, and $q = \zeta(3) \approx 1.2020569$. Therefore, $p - q \approx 0.4428$.

Now, let me compute the double sum numerically for a few terms. Let me compute the first few terms of the original double sum.

For $j = 1, k = 1$: $1/(1+1)^3 = 1/8 = 0.125$

For $j = 1, k = 2$: $1/(3)^3 = 1/27 \approx 0.037037$

For $j = 1, k = 3$: $1/4^3 = 1/64 \approx 0.015625$

Similarly for $j = 2, k = 1$: same as j=1,k=2, so another 0.037037

j=2,k=2: $1/4^3 = 0.015625$

j=2,k=3: $1/5^3 = 0.008$

j=3,k=1: same as j=1,k=3: 0.015625

j=3,k=2: same as j=2,k=3: 0.008

j=3,k=3: $1/6^3 \approx 0.0046296$

So adding up the first few terms:

For n=2: 1 term: 1/8 = 0.125

n=3: 2 terms: $2*(1/27) \approx 0.074074$

n=4: 3 terms: $3*(1/64) \approx 0.046875$

n=5: 4 terms: 4*(1/125) = 0.032

n=6: 5 terms: $5*(1/216) \approx 0.023148$

n=7: 6 terms: $6*(1/343) \approx 0.0175$

n=8: 7 terms: $7*(1/512) \approx 0.01367$

Adding these up:

$0.125 + 0.074074 \approx 0.199074$

$+ 0.046875 \approx 0.245949$

$+ 0.032 \approx 0.277949$

$+ 0.023148 \approx 0.301097$

$+ 0.0175 \approx 0.318597$

$+ 0.01367 \approx 0.332267$

Continuing this, the partial sums would approach approximately 0.4428 as per p - q. Let me check the value of p - q: $pi^2$ / 6 - zeta(3) $\approx 1.6449 - 1.2020569 \approx 0.4428$. So if we keep adding more terms, the partial sums should approach that value. So that seems plausible.

Alternatively, maybe there's another way to approach this problem. Let me think again.

Another approach: Maybe express the double sum as an integral? Since $1/n^3$ can be represented as an integral. Wait, but maybe that's more complicated.

Alternatively, note that the original double sum can be written as:

$$
\sum_{j=1}^\infty \sum_{k=1}^\infty \frac{1}{(j + k)^3}
$$

Let me consider switching the order of summation or using some other technique. Alternatively, maybe express it as:

$$
\sum_{j=1}^\infty \sum_{k=1}^\infty \frac{1}{(j + k)^3} = \sum_{j=1}^\infty \sum_{k=1}^\infty \int_0^1 x^{j + k - 1} dx \cdot \text{something?}
$$

Wait, perhaps integrating. Since $\frac{1}{n^3} = \int_0^1 \int_0^1 \int_0^1 x^{n - 1} y^{n - 1} z^{n - 1} dx dy dz$? Not sure. Alternatively, maybe using the integral representation of $1/n^3$. Wait, for example, we know that $\frac{1}{n^3} = \int_0^1 \int_0^1 \int_0^1 x^{n - 1} y^{n - 1} z^{n - 1} dx dy dz$, but that seems complicated.

Alternatively, maybe using the fact that $\frac{1}{n^3} = \int_0^1 \frac{x^{n - 1}}{2} dx$? Wait, no. Wait, integrating something else.

Alternatively, maybe using the integral representation of $1/n^3$. For example, we know that:

$$
\int_0^1 x^{n - 1} dx = \frac{1}{n}
$$

Then integrating again:

$$
\int_0^1 \int_0^1 x^{n - 1} y^{n - 1} dx dy = \frac{1}{n^2}
$$

Then integrating once more:

$$
\int_0^1 \int_0^1 \int_0^1 x^{n - 1} y^{n - 1} z^{n - 1} dx dy dz = \frac{1}{n^3}
$$

So, perhaps we can write:

\[
\begin{aligned}
\sum_{j=1}^\infty \sum_{k=1}^\infty \frac{1}{(j + k)^3} 
&= \sum_{j=1}^\infty \sum_{k=1}^\infty 
\int_0^1 \int_0^1 \int_0^1 x^{j + k - 1} y^{j + k - 1}\\
& z^{j + k - 1} dxdydz
\end{aligned}
\]

But this seems complicated. Maybe switching the order of summation and integration? If I can interchange the sum and the integral, then:

$$
\int_0^1 \int_0^1 \int_0^1 \sum_{j=1}^\infty \sum_{k=1}^\infty (xyz)^{j + k - 1} dx dy dz
$$

But let me compute the inner sum. Let me denote $a = xyz$. Then the inner sum is:

$$
\sum_{j=1}^\infty \sum_{k=1}^\infty a^{j + k - 1} = \sum_{j=1}^\infty a^{j - 1} \sum_{k=1}^\infty a^{k - 1} = \left( \sum_{m=0}^\infty a^m \right )^2 - 1 
$$
? Wait, no.

Wait, actually:

$$
\sum_{j=1}^\infty \sum_{k=1}^\infty a^{j + k - 1} = \sum_{j=1}^\infty a^{j - 1} \sum_{k=1}^\infty a^{k - 1} = \left( \sum_{m=0}^\infty a^m \right )^2 - 1 ?
$$

Wait, no. Wait, $\sum_{j=1}^\infty a^{j - 1} = \sum_{m=0}^\infty a^m = \frac{1}{1 - a}$, assuming |a| < 1. Similarly for the k sum. Therefore, the double sum is $\left( \frac{1}{1 - a} \right )^2$. But then:

$$
\sum_{j=1}^\infty \sum_{k=1}^\infty a^{j + k - 1} = \frac{1}{(1 - a)^2}
$$

But then the original expression is:

$$
\int_0^1 \int_0^1 \int_0^1 \frac{1}{(1 - xyz)^2} dx dy dz
$$

But this seems more complicated than the previous approach. Also, I don't know if this integral can be evaluated easily. Maybe this approach is not helpful here. Let me think again.

Alternatively, maybe using generating functions or other techniques. However, since my first approach gave me p - q, and the numerical check seems to support that, maybe that is the correct answer.

But let me check again. The original double sum is equal to $\sum_{n=2}^\infty \frac{n - 1}{n^3}$. Then splitting into $\sum_{n=2}^\infty \frac{1}{n^2} - \sum_{n=2}^\infty \frac{1}{n^3}$. Which is equal to $p - 1 - (q - 1) = p - q$. Therefore, the answer is p - q.

But wait, let me check once more. Suppose I take the original double sum:

$$
\sum_{j=1}^\infty \sum_{k=1}^\infty \frac{1}{(j + k)^3}
$$

If I let m = j + k, then for each $m \geq 2$, the number of pairs (j, k) is m - 1, so the sum becomes:

\[
\begin{aligned}
\sum_{m=2}^\infty \frac{m - 1}{m^3} &= \sum_{m=2}^\infty \left( \frac{1}{m^2} - \frac{1}{m^3} \right ) = \left( \sum_{m=1}^\infty \frac{1}{m^2} - 1 \right ) - \\&\left( \sum_{m=1}^\infty \frac{1}{m^3} - 1 \right ) = p - 1 - q + 1 = p - q
\end{aligned}
\]

Yes, that seems correct. Therefore, the answer should be $p - q$. But let me check with another approach to be sure.

Another way: Let me consider that the double sum can be written as:

$$
\sum_{j=1}^\infty \sum_{k=1}^\infty \frac{1}{(j + k)^3} = \sum_{j=1}^\infty \sum_{k=1}^\infty \frac{1}{(j + k)^3}
$$

Let me fix j and sum over k. For each fixed j, let me set n = j + k, so k = n - j. Since $k \geq 1$, then $n \geq j + 1$. Therefore, for each j, the inner sum over k is:

$$
\sum_{k=1}^\infty \frac{1}{(j + k)^3} = \sum_{n = j + 1}^\infty \frac{1}{n^3} = \sum_{n=1}^\infty \frac{1}{n^3} - \sum_{n=1}^j \frac{1}{n^3} = q - \sum_{n=1}^j \frac{1}{n^3}
$$

Therefore, the entire double sum is:

$$
\sum_{j=1}^\infty \left( q - \sum_{n=1}^j \frac{1}{n^3} \right ) = \sum_{j=1}^\infty q - \sum_{j=1}^\infty \sum_{n=1}^j \frac{1}{n^3}
$$

But the first term is $q \sum_{j=1}^\infty 1$, which diverges. Wait, that can't be. But this contradicts our previous conclusion. What's wrong here?

Ah! Wait, no. Because in reality, the original double sum is convergent? Wait, but if I do this approach, I get a divergent series. Which suggests that there's a mistake in this approach. But earlier, we had a convergent result. Therefore, there must be a mistake in this reasoning.

Wait, let me check again. The original double sum is convergent? Let me check. The terms are $\frac{1}{(j + k)^3}$. For large j and k, the terms behave like $\frac{1}{(j + k)^3}$. The double sum over j and k would be similar to integrating over the first quadrant, so the sum should converge. For example, compare to the integral over x and y from 1 to infinity of $1/(x + y)^3 dx dy$. Which is convergent. So the original double sum is convergent.

But in the second approach, when I fixed j and summed over k, I get:

$$
\sum_{j=1}^\infty \left( q - \sum_{n=1}^j \frac{1}{n^3} \right )
$$

But this is equal to $\sum_{j=1}^\infty q - \sum_{j=1}^\infty \sum_{n=1}^j \frac{1}{n^3}$. However, the first term is $\infty \cdot q$, which diverges. But that can't be. Therefore, there must be a mistake in this approach. Wait, but why?

Ah! Because when I fixed j and summed over k, the inner sum over k is from k=1 to infinity, but for each j, the inner sum is $\sum_{k=1}^\infty \frac{1}{(j + k)^3}$. However, when I express this as $\sum_{n = j + 1}^\infty \frac{1}{n^3}$, which is $q - \sum_{n=1}^j \frac{1}{n^3}$. However, when I then sum over j from 1 to infinity, the problem is that $q$ is a constant, so $\sum_{j=1}^\infty q$ is divergent, but the original double sum is convergent. Therefore, this suggests that there's a mistake in this approach. But where?

Wait, maybe I made a mistake in the transformation. Let me check again. The inner sum over k is:

\[
\begin{aligned}
\sum_{k=1}^\infty \frac{1}{(j + k)^3} &= \sum_{m = j + 1}^\infty \frac{1}{m^3} = \sum_{m=1}^\infty \frac{1}{m^3} - \sum_{m=1}^j \frac{1}{m^3} = \\&q - \sum_{m=1}^j \frac{1}{m^3}
\end{aligned}
\]

Yes, that's correct. But then summing over j from 1 to infinity:

$$
\sum_{j=1}^\infty \left( q - \sum_{m=1}^j \frac{1}{m^3} \right ) = \sum_{j=1}^\infty q - \sum_{j=1}^\infty \sum_{m=1}^j \frac{1}{m^3}
$$

But the first term is divergent. However, the original double sum is convergent. Therefore, this suggests that there is an error in the reasoning here. However, the first approach gave us a finite answer. Therefore, there must be something wrong with this second approach.

Wait, but maybe the problem is that when I write the inner sum as $q - \sum_{m=1}^j \frac{1}{m^3}$, but actually, q is the sum from m=1 to infinity, so if we subtract the sum up to j, that's correct. However, when we sum over j, the problem is that the term $q$ is infinite, but the original sum is finite. Therefore, this suggests that the two expressions must be equal, but one is divergent and the other is convergent. Therefore, this implies that there is a mistake in my reasoning here. Alternatively, maybe the mistake is that I cannot interchange the order of summation? Wait, but the original double sum is convergent, so interchanging the order should be allowed? Wait, but in this case, when I fix j and sum over k, then sum over j, I get something divergent, but the original is convergent. Therefore, there must be a mistake.

Wait, let me check with specific numbers. Let me compute the inner sum for a fixed j. For example, take j=1:

Inner sum over k: $\sum_{k=1}^\infty \frac{1}{(1 + k)^3} = \sum_{m=2}^\infty \frac{1}{m^3} = q - 1$

Similarly, for j=2: $\sum_{k=1}^\infty \frac{1}{(2 + k)^3} = \sum_{m=3}^\infty \frac{1}{m^3} = q - 1 - \frac{1}{2^3}$

Therefore, the total double sum is:

$$
(q - 1) + (q - 1 - \frac{1}{8}) + (q - 1 - \frac{1}{8} - \frac{1}{27}) + \cdots
$$

Which is:

$$
\sum_{j=1}^\infty \left( q - \sum_{m=1}^j \frac{1}{m^3} \right ) = \sum_{j=1}^\infty q - \sum_{j=1}^\infty \sum_{m=1}^j \frac{1}{m^3}
$$

But the first term is $\infty \cdot q$, which is divergent, but the original double sum is convergent. Therefore, this suggests that there is a problem here. However, this must mean that my initial approach was wrong? But the first approach gave a finite answer. Where is the mistake?

Wait, but if I write the original double sum as $\sum_{n=2}^\infty \frac{n - 1}{n^3}$, which is convergent, and that equals p - q, then that must be correct. However, the second approach seems to lead to a contradiction. Therefore, there must be a mistake in the second approach.

Wait, perhaps I made a mistake in the second approach. Let me think again. The problem is that when I fix j and sum over k, I get an expression that is q minus something, but when I sum over j, the terms are not independent. However, the original double sum is convergent, so the two expressions must be equal. Therefore, the problem is that the second approach is not valid because the sum over j of (q - something) is divergent, but the original sum is convergent. Therefore, there must be some cancellation between the divergent terms. However, how?

Wait, but if I write the double sum as:

$$
\sum_{j=1}^\infty \sum_{k=1}^\infty \frac{1}{(j + k)^3} = \sum_{j=1}^\infty \left( q - \sum_{m=1}^j \frac{1}{m^3} \right )
$$

But this is equal to:

$$
\sum_{j=1}^\infty q - \sum_{j=1}^\infty \sum_{m=1}^j \frac{1}{m^3}
$$

But if I denote $S = \sum_{j=1}^\infty \sum_{m=1}^j \frac{1}{m^3}$, then the original sum is $\infty \cdot q - S$. But this is problematic. However, if I consider that the original sum is convergent, then this suggests that $\infty \cdot q - S$ is convergent, which is only possible if S is divergent to infinity as well, but with the same rate as q. But this seems complicated.

Alternatively, perhaps the second approach is not valid because the inner sum is not equal to $q - sum_{m=1}^j 1/m^3$? Wait, no, that is correct. For example, when j=1, the inner sum is q - 1. When j=2, it's q - 1 - 1/8. When j=3, it's q - 1 - 1/8 - 1/27, etc. Therefore, the total sum is:

$$
(q - 1) + (q - 1 - 1/8) + (q - 1 - 1/8 - 1/27) + \cdots
$$

Which can be rewritten as:

$$
\sum_{j=1}^\infty (q - 1) - \sum_{j=1}^\infty \left( \sum_{m=2}^j \frac{1}{m^3} \right )
$$

Wait, no, more accurately:

Each term is q minus the sum from m=1 to j of $1/m^3$. Therefore, the total sum is:

$$
\sum_{j=1}^\infty q - \sum_{j=1}^\infty \sum_{m=1}^j \frac{1}{m^3}
$$

But this is equal to:

$$
\sum_{j=1}^\infty q - \sum_{m=1}^\infty \frac{1}{m^3} \sum_{j=m}^\infty 1
$$

Wait, no. Wait, the double sum $\sum_{j=1}^\infty \sum_{m=1}^j \frac{1}{m^3}$ is equal to $\sum_{m=1}^\infty \frac{1}{m^3} \sum_{j=m}^\infty 1$, which is divergent because $\sum_{j=m}^\infty 1$ is infinite. Therefore, this suggests that the second approach is leading to a divergent expression, which contradicts the first approach. Therefore, there must be a mistake in the second approach.

But why is that? The original double sum is convergent, but when we try to express it in terms of q and other sums, we get a divergent expression. This suggests that there is an error in the second approach. However, the first approach is straightforward and gives a finite answer. Therefore, the second approach must be invalid. But why?

Ah! Wait, I think the problem is that when I fixed j and summed over k, I assumed that the inner sum is $q - sum_{m=1}^j 1/m^3$, but actually, q is an infinite sum, so when I subtract the finite sum up to j, it's okay. However, when I then sum over j, I get an infinite number of terms each of which is q minus something. However, the original double sum is convergent, so this suggests that the divergent parts must cancel out. However, in reality, this is not the case. Therefore, there must be a mistake in the second approach.

Alternatively, perhaps the problem is that the original double sum is not equal to the expression I derived in the second approach? But how?

Wait, let me check with specific values. Let me compute the original double sum up to j=1 and k=1, which is 1/8. According to the second approach, for j=1, the inner sum is q - 1. But q is approximately 1.202, so $q - 1 \approx 0.202$, which is larger than the actual value of 1/8 = 0.125. Therefore, this suggests that the second approach is wrong. Therefore, there is a mistake in the second approach.

Wait, but earlier I thought that the inner sum for j=1 is $sum_{k=1}^\infty 1/(1 + k)^3 = sum_{m=2}^\infty 1/m^3 = q - 1$. However, if q is approximately 1.202, then $q - 1 \approx 0.202$. But the actual value of the inner sum is $sum_{m=2}^\infty 1/m^3 \approx 0.202$, which is correct. However, when we take the first term of the double sum (j=1, k=1) is 1/8 = 0.125, but the inner sum for j=1 is approximately 0.202. Therefore, when we take the double sum over j and k, it's not just the first term, but all terms. However, when I compute the inner sum for j=1, it's 0.202, which includes the term 1/8, 1/27, 1/64, etc. So when I sum over j=1 to infinity, I get the total sum. However, when I write the total sum as $sum_{j=1}^\infty (q - sum_{m=1}^j 1/m^3)$, which is divergent, but the original sum is convergent. Therefore, this suggests that there is a mistake in the second approach.

But why is that? What is the problem here? It seems that the second approach is not valid. However, the first approach is correct. Therefore, I must have made a mistake in the second approach.

Alternatively, maybe I need to consider that the original double sum is equal to the first approach's answer, which is p - q, and the second approach is wrong. Therefore, the correct answer is p - q. Therefore, maybe the second approach is invalid because of some misapplication of summation.

Alternatively, maybe the problem is that the second approach is not valid because when I fix j and sum over k, I have to consider that the inner sum is $q - sum_{m=1}^j 1/m^3$, but when I sum over j, I need to consider that the sum is over j from 1 to infinity, but q is a constant. However, the original double sum is convergent, so this suggests that the divergent terms must cancel. However, this seems impossible unless there's a miscalculation.

Alternatively, maybe I need to think about the difference between the two approaches. The first approach gives a finite answer, the second approach seems to give a divergent expression, but the original sum is convergent. Therefore, the first approach is correct, and the second approach is wrong. Therefore, the answer is p - q.

Alternatively, maybe I can check with another method. Let me consider that the original double sum can be written as:

$$
\sum_{j=1}^\infty \sum_{k=1}^\infty \frac{1}{(j + k)^3} = \sum_{n=2}^\infty \frac{n - 1}{n^3}
$$

Which is equal to $\sum_{n=2}^\infty \frac{1}{n^2} - \sum_{n=2}^\infty \frac{1}{n^3} = (p - 1) - (q - 1) = p - q$. Therefore, this is correct.

Therefore, the second approach must have a mistake. Perhaps when I fixed j and summed over k, I made an error in the interpretation. However, since the first approach is straightforward and the numerical check supports it, I think that the correct answer is p - q.

Therefore, the answer is $\boxed{p - q}$.

**Final Answer**
\boxed{p - q}

\textless{}/think\textgreater{}

To evaluate the double sum

$$
\sum_{j = 1}^\infty \sum_{k = 1}^\infty \frac{1}{(j + k)^3},
$$

we aim to express it in terms of the known constants:

$$
p = \sum_{k = 1}^\infty \frac{1}{k^2}, \quad q = \sum_{k = 1}^\infty \frac{1}{k^3}.
$$

---
Step 1: Change of Variables

We define a new variable $ n = j + k $. Since both $ j $ and $ k $ are positive integers starting at 1, the smallest possible value of $ n $ is $ 2 $ (when $ j = k = 1 $).

For each fixed $ n \geq 2 $, the number of pairs $ (j, k) $ such that $ j + k = n $ is $ n - 1 $. For example:

- $ n = 2 $: $ (1, 1) $ → 1 pair
- $ n = 3 $: $ (1, 2), (2, 1) $ → 2 pairs
- $ n = 4 $: $ (1, 3), (2, 2), (3, 1) $ → 3 pairs

Hence, the double sum can be rewritten as:

$$
\sum_{j = 1}^\infty \sum_{k = 1}^\infty \frac{1}{(j + k)^3} = \sum_{n = 2}^\infty \frac{n - 1}{n^3}.
$$

---
Step 2: Split the Sum

We now split the sum:

$$
\sum_{n = 2}^\infty \frac{n - 1}{n^3} = \sum_{n = 2}^\infty \left( \frac{1}{n^2} - \frac{1}{n^3} \right).
$$

This becomes:

$$
\left( \sum_{n = 2}^\infty \frac{1}{n^2} \right) - \left( \sum_{n = 2}^\infty \frac{1}{n^3} \right).
$$

---
Step 3: Express in Terms of $ p $ and $ q $

We know:

- $ p = \sum_{n = 1}^\infty \frac{1}{n^2} \Rightarrow \sum_{n = 2}^\infty \frac{1}{n^2} = p - 1 $
- $ q = \sum_{n = 1}^\infty \frac{1}{n^3} \Rightarrow \sum_{n = 2}^\infty \frac{1}{n^3} = q - 1 $

Substitute these into the expression:

$$
(p - 1) - (q - 1) = p - q.
$$

---
Final Answer

$$
\boxed{p - q}
$$
\end{CASE}

\end{document}